\pdfoutput=1

\documentclass[11pt]{article}

\usepackage[preprint]{acl}

\usepackage{times}
\usepackage{float}
\usepackage{latexsym}
\usepackage{graphicx} 
\usepackage{booktabs} 
\usepackage{tcolorbox}
\usepackage{hyperref}
\usepackage{multirow} 
\usepackage{adjustbox}
\usepackage{amsmath}
\usepackage{inconsolata}
\usepackage{xcolor} 
\usepackage{pdflscape}  
\usepackage{lscape}    
\usepackage{colortbl}
\usepackage{longtable}
\usepackage{array}
\usepackage{enumitem}
\usepackage{todonotes}
\newcounter{todocounter}
\usepackage{microtype}
\usepackage{tabularx}
\usepackage{rotating}
\usepackage{xspace,mfirstuc,tabulary}
\usepackage{rotating}

\usepackage[T1]{fontenc}

\usepackage[utf8]{inputenc}

\usepackage{microtype}

\usepackage{inconsolata}

\usepackage{graphicx}

\newtcolorbox[auto counter, number within=section]{myboxA}[3][]{%
  colback=teal!10!white,
  colframe=teal!40!black,
  fonttitle=\bfseries,
  title=#3,
  boxsep=2pt,
  left=3pt,
  right=3pt,
  top=3pt,
  bottom=3pt,
  toptitle=2pt,
  bottomtitle=2pt,
  fontupper=\footnotesize,
  #1
}

\newtcolorbox[auto counter, number within=section]{myboxB}[3][]{%
  colback=blue!10!white,
  colframe=blue!40!black,
  fonttitle=\bfseries,
  title=#3,
  boxsep=2pt,
  left=3pt,
  right=3pt,
  top=3pt,
  bottom=3pt,
  toptitle=2pt,
  bottomtitle=2pt,
  fontupper=\footnotesize,
  #1
}

\title{Bridging the Gap: In-Context Learning for Modeling Human Disagreement}

\author{
  \textbf{Benedetta Muscato\textsuperscript{1}},
\textbf{Yue Li\textsuperscript{2}},
\textbf{Gizem Gezici \textsuperscript{1}},
\textbf{ Zhixue Zhao \textsuperscript{2}}, \textbf{Fosca Giannotti \textsuperscript{1}}
\\
\\
\textsuperscript{1}Scuola Normale Superiore, Pisa, Italy
\\
\textsuperscript{2}University of Sheffield, Sheffield, UK
\\
\small{
    \textbf{Correspondence: Benedetta Muscato, } \href{mailto:email@domain}{benedetta.muscato@sns.it}
  }
}

\begin{document}
\maketitle
\begin{abstract}
Large Language Models (LLMs) have shown strong performance on NLP classification tasks. However, they typically rely on aggregated labels—often via majority voting—which can obscure the human disagreement inherent in subjective annotations.
This study examines whether LLMs can capture multiple perspectives and reflect annotator disagreement in subjective tasks such as hate speech and offensive language detection. We use in-context learning (ICL) in zero-shot and few-shot settings, evaluating four open-source LLMs across three label modeling strategies: aggregated hard labels, and disaggregated hard and soft labels.
In few-shot prompting, we assess demonstration selection methods based on textual similarity (BM25, PLM-based), annotation disagreement (entropy), a combined ranking, and example ordering strategies (random vs. curriculum-based).
Results show that multi-perspective generation is viable in zero-shot settings, while few-shot setups often fail to capture the full spectrum of human judgments. Prompt design and demonstration selection notably affect performance, though example ordering has limited impact.
These findings highlight the challenges of modeling subjectivity with LLMs and the importance of building more perspective-aware, socially intelligent models.

\end{abstract}

\noindent \textbf{Warning:} \textit{{This paper contains examples that may be offensive or upsetting.}}

\begin{figure}[!t]
    \centering
 
    \includegraphics[width=0.75\linewidth]{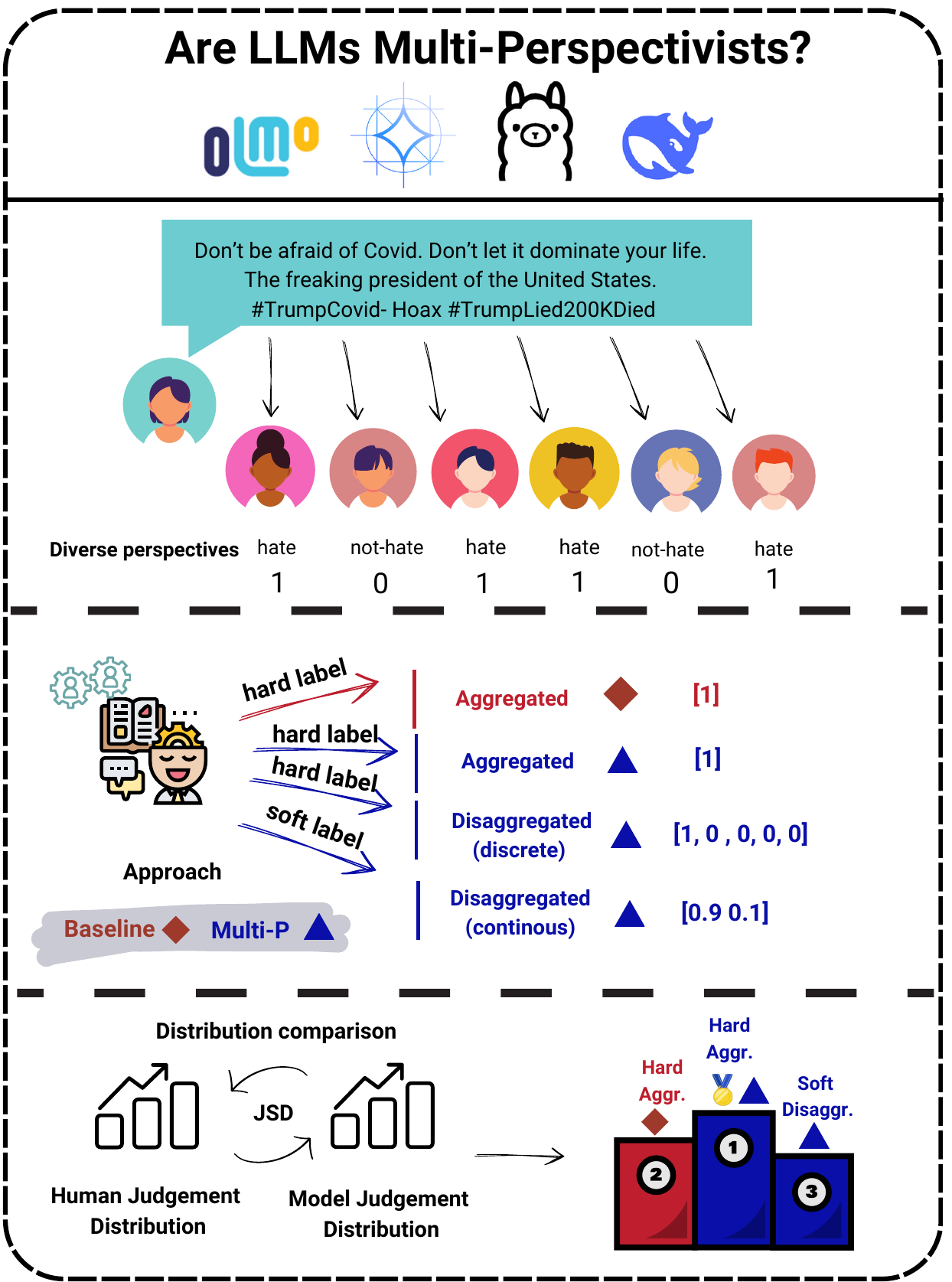} 

    \vspace{-0.7em}
    \caption{Overview of our findings. Our analysis shows that LLMs can effectively reflect multiple perspectives in hate speech detection, when prompted to consider diverse viewpoints or when predicting aggregated labels. This is illustrated using Jensen-Shannon Divergence (JSD) on a representative annotation example.
    }
    \label{fig:findings}
    \vspace{-1.2em}  
\end{figure}

\section{Introduction}
\label{sec:intro}
In traditional classification tasks, annotations are typically collected from multiple annotators, often through crowdsourcing, and consolidated into a \textit{single} ground truth label per example.
While this method works well when annotators are in full agreement and the aggregated label reflects a clear consensus, it fails to account for cases where multiple, equally valid interpretations are possible.
This shortcoming is especially pronounced in subjective tasks\footnote{In Natural Language Processing (NLP), subjective tasks are those with multiple valid interpretations, influenced by personal perspectives, emotions, or social context~\cite{rottger2022two}.}
where inherent annotator disagreement can emerge for valid reasons such as text ambiguity, genuine disagreement, subjective preferences, and two or more plausible perspectives \cite{plank-etal-2014-linguistically}.
For instance, hate speech detection often entails contentious annotations, as individuals may interpret hateful content differently based on their personal experiences or cultural backgrounds \cite{davani2021hatespeechclassifierslearn, akhtar2021whose}. 
Similarly, the detection of offensive and abusive language is highly dependent on subjective judgments, making the task even more complex ~\cite{del2021offendes,husain2021survey}.

A new line of research highlights that leveraging disagreements in human-annotated datasets can enhance both model performance and confidence  \cite{casola2023confidence, davani2022dealing, sandri2023don,muscato2024multi}.
This emerging direction, known as Perspectivism \cite{basile-etal-2021-need}, suggests a paradigm shift in model design philosophy, advocating for the development of systems that are not only \textit{Perspective-aware} but also more \textit{Responsible} and \textit{Socially Intelligent} \cite{liu2023training, kovavc2023large}.

Nevertheless, LLMs, including systems such as GPT-series, exhibit notable limitations in modeling the diversity of user preferences \cite{pavlovic-poesio-2024-effectiveness,feng2024modular} across different sociodemographic groups \cite{wang2025large}.
~\citet{srivastava2022beyond} show that LLMs are prone to inherent biases, which are particularly pronounced in ambiguous contexts where human judgments are subjective.
Similarly,~\citet{santurkar2023whose} highlight that LLMs tend to reflect a predominantly left-leaning perspective, further limiting their ability to provide diverse opinions.

In this paper, we focus on the interplay between in-context learning (ICL) and human disagreement in subjective NLP tasks. Specifically, we investigate \textbf{whether multi-perspective ICL can serve as an effective and practical alternative to fine-tuning for capturing diverse perspectives}, especially with limited annotated data and computational resources.  
Through a comprehensive analysis of established ICL methods, we evaluate their performance in both zero-shot and few-shot prompting.
Our study explores thirty different combinations across~\emph{four LLMs}, ~\emph{three subjective tasks} (hate speech, offensive language, and abusive language detection) in Section~\ref{sec:task_definition},~\emph{three types of label spaces} (aggregated hard and disaggregated hard \& soft) in Section~\ref{sec:label_space}, and~\emph{five demonstration example strategies} in Section~\ref{sec:demonstration_examples}. These include both~\emph{selection methods} (textual similarity, annotator disagreement, and two-stage ranking) and~\emph{ordering techniques} (random and In-Context Curriculum Learning (ICCL)~\cite{liu2024let}).
Our multi-perspective ICL approach, which specifically instructs the model to consider different viewpoints, is compared to the standard prompting (baseline).
Although our findings demonstrate that the multi-perspective approach performs better than the baseline in predicting aggregated labels, it struggles to capture subjective nuances in disaggregated labels. 
Our findings highlight the need for perspective-aware LLMs to mitigate polarization and societal bias, as shown in Figure~\ref{fig:findings}, and underscores ongoing challenges in applying ICL to subjective tasks.

The closest work to our methodology is~\citet{pavlovic-poesio-2024-effectiveness}, which uses a single closed-model LLM (GPT-3.5-turbo) to compare human and LLM opinion distributions leveraging zero-shot ICL with role-playing, generating only disaggregated soft labels in both our English benchmark datasets, along with an additional Arabic one. 
We extend prior work in the following ways: 1) using four open-source LLMs instead of a single closed model, 2) prompting the models in both zero-shot and few-shot settings, 3) incorporating tailored selection and ordering strategies for demonstration organization in few-shot prompting, specifically designed for subjective tasks and 4) expanding the label space with: aggregated hard labels, disaggregated hard \& soft labels.
Thus, we posit that our results provide generalizable evidence on the limitations and potential of LLMs to handle subjective disagreement through ICL.

\section{Background}
\subsection{Socially Intelligent and Responsible LLMs}

An AI system is deemed socially intelligent when it can understand and reason about the intentions, emotions, and mental states of others~\cite{sap2022neural, pereira2016integrating}.
Beyond basic functionality, these models or agents aim to sense, perceive, reason about, learn from, and respond to the affective behavior and cognition of other agents (whether human or artificial)~\cite{qiu-etal-2022-towards}.
Like humans, they evolve their perspectives through interaction, integrating subjective preferences and beliefs~\cite{mittelstadt2024large, kirk2025human}.
In the broader context of Responsible AI, which aims not just at better performance, but at delivering societal benefit, social intelligence helps align LLMs with diverse user preferences~\cite{mathur2024advancing}, reducing harm and fostering pluralistic alignment across individual and collective needs~\cite{sorensen2024position, tahaei2023human, kirk2024benefits}.

\subsection{Learning from Human Disagreement}
\label{sec:disagreement}

Human disagreement, often seen as noise that harms label quality in crowd-sourcing \cite{artstein2017inter, fleisig2024perspectivist}, is increasingly reframed as 
a form of plausible~\emph{human label variation} (HLV)~\cite{plank2022problem}, especially in subjective tasks e.g., hate speech, stance, and emotion detection, where no single ground truth label may exist. 
While traditional methods rely on hard (discrete) aggregated labels, such as those consolidated through majority voting, the perspectivist literature~\cite{frenda2024perspectivist} preserves disagreement using either hard or soft (continuous) disaggregated labels. Recent research shows that models fine-tuned on disaggregated soft labels outperform those using aggregated hard labels~\cite{van2024annotator,uma2021learning}.
Alternative approaches include training an ensemble of classifiers, one for each annotator~\cite{akhtar2021whose}, adopting single-task and multi-task architectures~\cite{davani2022dealing}, and incorporating socio-demographic information from target groups~\cite{fleisig2023majority}. For a comprehensive review of fine-tuning with human disagreement, refer to~\cite{uma2021learning} and \cite{frenda2024perspectivist}.

\subsection{In-Context Learning (ICL)}

ICL enables LLMs to learn new tasks by analogy without parameter updates~\cite{dong2024survey, winston1980learning}.
While zero-shot uses no examples, few-shot provides a small demonstration set. ICL excels in complex reasoning~\cite{wei2022chain} and role-playing~\cite{kong2024self}, but faces some limitations: high computational cost, lower efficiency than PEFT~\cite{liu2022few}, and sensitivity to prompt design, i.e. example selection~\cite{gao2021making} and ordering~\cite{liu2024let, peng-etal-2024-revisiting}.
While cost-effective, ICL’s potential in subjective tasks remains underexplored. \citet{chen-etal-2024-seeing} show LLMs can approximate human label distributions from a few expert-labeled examples with explanations, but it is unclear if this holds for non-experts, whose disagreements do not imply low-quality annotations.
Furthermore, prompt sensitivity remains a key challenge, reducing their reliability in broader applications, especially when subjective preferences are involved.

\paragraph{Demonstration Selection and Ordering}
ICL performance is highly sensitive to how demonstration examples are selected and ordered. Traditional selection methods, such as random sampling~\cite{brown2020language} or human-curated examples~\cite{kazemi2023lambada}, yield mixed results.
Advanced methods include KATE, a kNN-based selection approach~\cite{liu2022makes} and a latent token-based method~\cite{wang2024large}, improve ICL by selecting semantically similar demonstration examples that are likely to predict conceptual tokens. Other strategies utilize mutual information~\cite{sorensen2022information} or perplexity~\cite{gonen-etal-2023-demystifying} to refine the selection process.
Demonstration ordering is also a critical factor; mitigation strategies include iterative selection with chain-of-thought reasoning~\cite{qin2024context}, and entropy-based ordering~\cite{lu-etal-2022-fantastically}.

\section{Prompting LLMs for Multi-Perspective}
\label{sec:ICL}

This section presents our method for assessing whether ICL enables LLMs to account for diverse perspectives in subjective tasks by prompting them with explicit multi-perspective instructions\footnote{Further details on the prompts can be found at~\url{https://anonymous.4open.science/r/Multi_Perspective_ICL-3203}. The code will be shared upon acceptance.}.
Formally, let $M$ be the model (LLM) and $t$ the subjective task, defined by input text $x$ and a set of annotations A$ = {a_1, ..., a_n}$ with $n$ annotators.
The model $M$ is then prompted to predict the label of $x$, denoted as $\hat{y}$, using zero-shot or few-shot ICL with the following components~\cite{wang-etal-2022-super}:

\begin{itemize}
    \item \underline{Task Definition}: 
    A description of a subjective task $t$ explains how the input text $x$ should be analyzed in order to be mapped to the appropriate label space.
    \item \underline{Label Space}: 
    A specification of the expected output label $\hat{y}$, whether as an aggregated or disaggregated hard or soft label $l$.
    \item \underline{Demonstration Examples}: 
   A reference input-output pairs $D = \{x_j', y_j'\}$, for $j=\{1,...,m\}$ (with $m$ examples), only for few-shot learning.
\end{itemize}

To investigate the multi-perspective capabilities of LLMs with ICL, we test various strategies for each prompt component: task definition (Section~\ref{sec:task_definition}), label space configuration (Section~\ref{sec:label_space}), and the arrangement of demonstration examples (Section~\ref{sec:demonstration_examples}).
Our multi-perspective (MP) prompt template is illustrated in Box~\ref{box:prompt_structure} $\uparrow$ in green.
Then, a sample prompt is shown in Box~\ref{box:prompt_multip} $\downarrow$ in purple, where $t$ contains the instructions for hate speech detection, and $l$ is an aggregated hard label. Since a demonstration example is provided to the model, this is a few-shot setting.

\subsection{Task Definition} 
\label{sec:task_definition}

In standard practice, LLMs are prompted to directly answer questions (e.g., Classify the following tweet as hate speech based on the options~\cite{antypas-etal-2023-supertweeteval}), without considering the inherent subjectivity of the task. In our study, we explore two approaches: the baseline (standard) and the multi-perspective approach.

\paragraph{Baseline Priming}
\label{par:baseline}

The baseline reflects a scenario where $M$~\emph{is prompted to produce a single aggregated label}, disregarding the subjectivity or ambiguity of $t$. 
Specifically, the example in Box~\ref{box:prompt_multip} $\downarrow$ in purple is adjusted, where $t$ does not contain the bold statement and $l$ remains the same to obtain $\hat{y}$ as an aggregated hard label.

\paragraph{Multi-Perspective Priming} 
\label{par:multip}

To reduce the tendency of $M$ to favor a dominant viewpoint and improve its contextual understanding of $t$, the model 
is~\emph{explicitly instructed to consider multiple perspectives} from different groups, as highlighted in the bold statement shown in Box~\ref{box:prompt_multip} $\downarrow$ in purple.
Inspired by \citet{pavlovic-poesio-2024-effectiveness,lan2024stance}, we prompt $M$ to~\emph{adopt the role of an expert in task $t$}. We apply role playing for both baseline and multi-perspective.

\begin{myboxA}[label=box:prompt_structure]{}{Our MP Prompt Template}{}
  \vspace{-0.5em}
    \textbf{\textsc{Task Definition ($t$):}} 
    \begin{itemize}[leftmargin=*,noitemsep,topsep=0pt]
        \item Hate speech
        \item Offensive language
        \item Abusive language
    \end{itemize}
    
    \textbf{\textsc{Label Space ($l$):}} 
    \begin{itemize}[leftmargin=*,noitemsep,topsep=0pt]
        \item \textbf{Hard}: Aggregated or Disaggregated
        \item \textbf{Soft} : Disaggregated
    \end{itemize}
    
    \textbf{\textsc{Demonstration Example(s) ($D$):}} 
    \begin{itemize}[leftmargin=*,noitemsep,topsep=0pt]
        \item \texttt{(text, hard agg.)}: (e.g., \texttt{yes})
        \item \texttt{(text, hard disagg.)}: (e.g., \texttt{[0,0,1,1,0]})
        \item \texttt{(text, soft)}: (e.g., \texttt{[0.7, 0.3]})
    \end{itemize}
    
    \textbf{\textsc{Input:}}  
    \begin{itemize}[leftmargin=*,noitemsep,topsep=0pt]
        \item Tweet ($x$): \texttt{\{text\}}
        \item Answer ($\hat{y}$): \texttt{[output]}
    \end{itemize}
    \vspace{-0.5em}
\end{myboxA}

\begin{myboxB}[label=box:prompt_multip]{}{Example MP Prompt for Hate Speech}{}
[$t$] Does the following tweet contain hate speech, particularly xenophobia or islamophobia? \textbf{The task is subjective, so please answer considering different perspectives} from Muslim immigrants as well as others from different backgrounds.
%

[$l$] There are two options: \textit{yes} and \textit{no}.

%
%
[$D$] Examples: Any future terrorist attack in Europe will be blame on Brexit by the lmsm, yes

Now consider the following example and only output your option without punctuation.
[$x$] Tweet: What the referendum seem to have mean to alarm number a vote for anyone look foreign to leave immediately

[$\hat{y}$] Answer:
\end{myboxB}

\subsection{Label Space} 
\label{sec:label_space}

To more effectively capture the subjective nature of the tasks, we define three types of label predictions with which  the multi-perspective approach prompts $M$ to generate: aggregated hard (discrete), disaggregated hard (discrete), and disaggregated soft labels (continuous).
If $M$ is prompted to generate $\hat{y}$ within the label space $l$, the demonstration examples are also represented within $l$.

\subsubsection{Hard Labels} 
\label{sec:hard}

\paragraph{Aggregated}

A single ground-truth, which is the final classification output (e.g., hate speech or not hate speech) derived from vote aggregation approaches such as majority voting.

\paragraph{Disaggregated} \label{par:disaggregated}
Individual labels from multiple annotators reflect the diversity of human judgments. Specifically, with a predefined number of annotators $n$, $M$ is prompted to generate labels in the form of $A'$, the set of individual annotations, where $a_i' \in A'$, for $i=\{1,...,n\}$, representing the predicted annotations from each annotator.

To convert aggregated and disaggregated hard labels into their disaggregated soft counterparts\footnote{This transformation is necessary only for the evaluation phase, for more details please refer to Section~\ref{subsec:eval_metrics}.}, we extract the probability scores from $M$ associated with the predicted labels~\cite{lee2023can}.

\begin{figure*}[!h]
    \centering
    \vspace{-2.0em}  
    \includegraphics[width=0.74\linewidth, keepaspectratio]{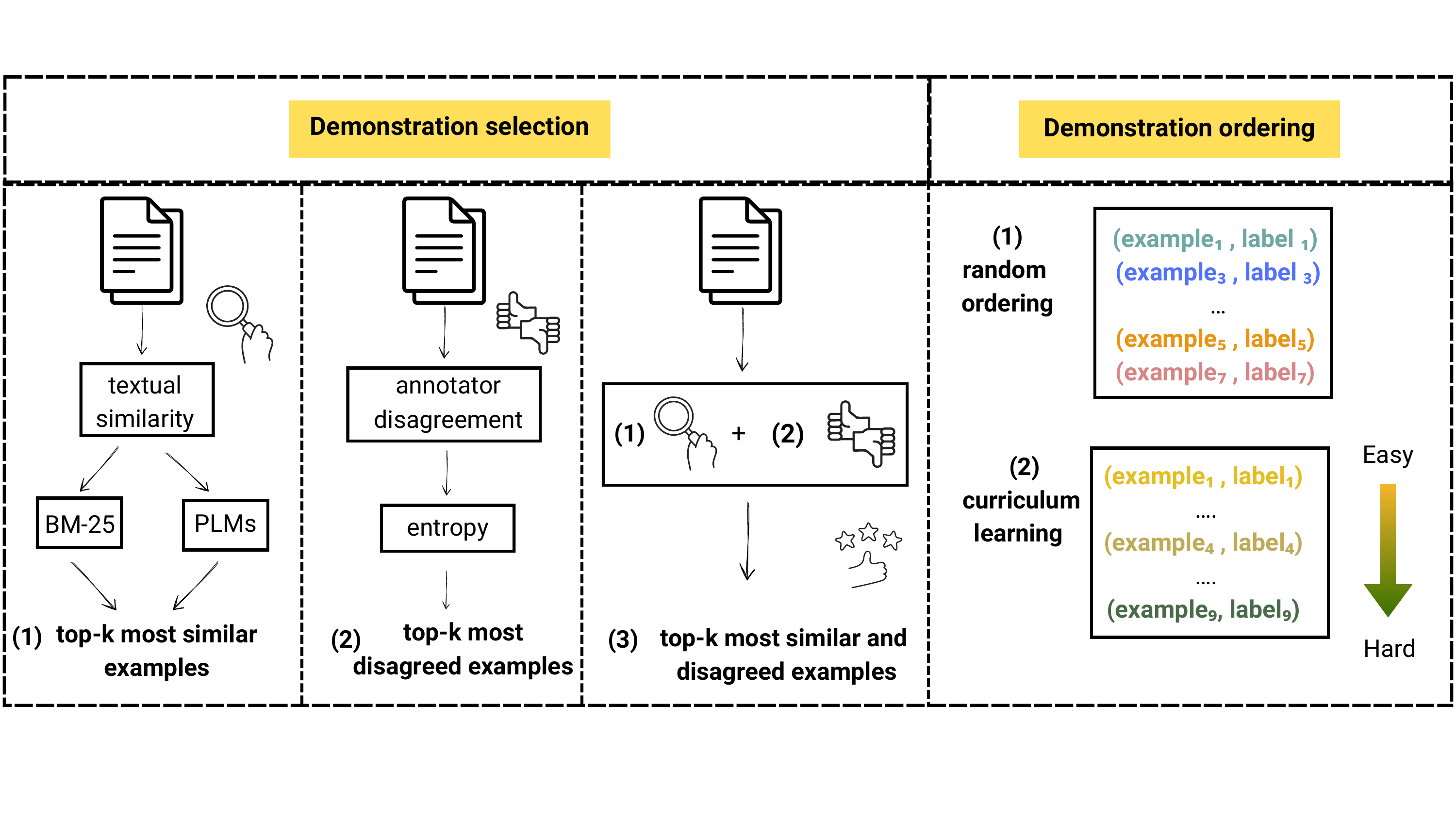}  
    \vspace{-2.9em}  
  \caption{We evaluate multi-perspective ICL by selecting demonstration examples through three strategies: (1) top-k most similar examples ranked via BM25 and pretrained language models (PLMs), (2) top-k examples with highest annotator disagreement measured by entropy, and (3) a two-stage method combining similarity retrieval followed by disagreement-based re-ranking. Additionally, we investigate the impact of two ordering strategies for the demonstrations on model performance.
  }
    \label{fig:fig2}
\vspace{-1em}  
\end{figure*}

\subsubsection{Soft Labels} 
\label{sec:soft}
\paragraph{Disaggregated} A probability distribution across the possible classes, indicating the likelihood of each category. For instance, if $t$ is a binary hate speech classification task, $M$ is prompted to generate probabilities in the format $[P_{hate}, P_{not\_hate}]$, ensuring that the values sum up to one. We observe that LLMs consistently adhere to this condition.

\subsection{Demonstration Examples}
\label{sec:demonstration_examples}

Prior research has made progress in optimizing demonstration organization~\cite{dong-etal-2024-survey}, but most studies focus on objective tasks. The effectiveness of these methods in subjective tasks, where LLMs must account for multiple perspectives, is under-explored. Furthermore, annotator disagreement, a key aspect of subjective tasks, has received little attention in example selection strategies. To address this, we explore different methods for~\emph{selecting} (Section~\ref{sec:selection}) and~\emph{ordering} (Section~\ref{sec:order}) demonstration examples in subjective tasks, as depicted in Figure~\ref{fig:fig2}, inspired by~\citet{liu2024let}.

\subsubsection{Demonstration Selection}\label{sec:selection}

We examine traditional approaches where textual similarity between the target input and examples plays a key role~\cite{peng-etal-2024-revisiting,zhang-etal-2024-teaching}. Given the subjective nature of the tasks, we also incorporate annotator disagreement into the example selection process. In total, we explore three approaches for k-shot example selection in few-shot ICL:

\paragraph{Textual Similarity} We calculate the similarity between the target test examples and the training set, then retrieve the top $k$ most similar texts from the training corpus based on a fixed threshold\footnote{We empirically tested various thresholds ranging from $0.5$ to $0.8$ and found that a threshold of $0.7$ produced the best performance in terms of cosine similarity.}. Similarity is measured using two different methods: (1) \textit{BM-25} \cite{robertson2009probabilistic}; and (2) following \citet{peng-etal-2024-revisiting}, cosine similarity between sentence embeddings obtained from \textit{Pre-trained Language Models (PLMs)}, namely all-MiniLM-L6-v2, all-MiniLM-L12-v2, all-distilroberta-v1, all-mpnet-base-v2 from huggingface\footnote{\url{https://huggingface.co/sentence-transformers}}.

\paragraph{Annotator Disagreement} We hypothesize that 
high annotator disagreement examples are more informative, as they reveal ambiguity or diverse interpretations of the subjective tasks. 
Thus, we select the top $k$  samples with the highest annotator disagreement (most disagreed) from the training set as the demonstration examples.
For each sample with $n$ annotations, we compute disagreement by calculating the entropy of the $n$ annotations. Higher entropy indicates greater disagreement, while lower entropy reflects higher agreement, therefore the instance is less difficult and more straightforward.

\paragraph{Two-Stage Ranking} Our preliminary experiments show that aforementioned methods may result in examples that are too similar or too diverse, hindering pattern learning.
To address this, we propose a two-stage approach inspired by information retrieval systems~\cite{dang2013two}, which combines both textual similarity and annotator disagreement. 
We first extract the top $k$ most similar training examples to the target test data. Then, we re-rank them by annotator disagreement, selecting the top $k$ examples that are both most similar and disagreed for few-shot ICL.

\subsubsection{Demonstration Ordering}
\label{sec:order}

We begin by using random ordering as a baseline and then explore an alternative strategy called In-Context Curriculum Learning (ICCL)~\cite{liu2024let}, which organizes examples in increasing order of difficulty. We hypothesize that the model can benefit from learning~\emph{simpler} examples first, those with high annotator~\emph{agreement}, before progressing to more~\emph{challenging} ones with higher {disagreement}. To measure difficulty, we use the entropy of annotations as a proxy and rank the selected $k$ examples from low to high entropy.

\section{Experimental Setup}

\subsection{Datasets}
\label{sec:datasets}
We use three benchmark datasets in English, each corresponding to a different subjective task $t$ from the SemEval 2023 (SE) \textit{Learning with Disagreements} (LeWiDi)\footnote{\url{https://le-wi-di.github.io}} competition \cite{leonardelli2023semeval}.
These datasets cover a range of subjective tasks with varying numbers of annotations $n$.
Our analysis focuses on the HS-Brexit, MD-Agreement, and ConvAbus datasets, with statistics shown in Table~\ref{tab:datasets}.

\begin{table}[!h]
    \centering
    \renewcommand{\arraystretch}{1.2}  
    \resizebox{0.48\textwidth}{!}{  
    \normalsize
    \begin{tabular}{lccccccc}  
        \toprule
        \textbf{Dataset} & \textbf{Train} & \textbf{Test} & \textbf{Dev} & \textbf{Tot. Class} & \textbf{Ann.} & \textbf{Full Agr. (\% )} & \textbf{Subj. Task} \\
        \midrule
        \textbf{HS-Brexit} & 784 & 168 & 168 & 2 & 6 & 69\% & Hate speech \\
        \textbf{MD-Agr} & 6592 & 3057 & 1104 & 2 & 5 & 42\% & Offensive lang. \\
        \textbf{ConvAbuse} & 2398 & 840 & 812 & 2 & 3-8 & 86\% & Abusive lang. \\
        \bottomrule
    \end{tabular}
    }
    \vspace{-0.9em}
    \caption{Dataset statistics from LeWiDi competition.}
    \label{tab:datasets}
    \vspace{-1.2em}
\end{table}

\paragraph{Hate Speech on Brexit}
\label{par:HS_brexit} 

HS-Brexit~\cite{akhtar2021whose} consists of English tweets collected before, during, and immediately after the Brexit vote. The tweets were filtered using keywords related to immigrants and Brexit to capture opinions about the role of immigrants in UK society. Six annotators were involved in the annotation process.

\paragraph{Multi-Domain Agreement}
\label{par:md_agr}

MD-Agreement \cite{leonardelli-etal-2021-agreeing} focuses on three different domains that were popular in online debates in 2020: Covid-19, the US Presidential elections and the Black Lives Matter (BLM) social movement. For each topic, a set of keywords is used to scrape related tweets. Once tweets are collected, each instance is labeled by five annotators.

\paragraph{Conv-Abuse}

ConvAbuse~\cite{cercas-curry-etal-2021-convabuse} consists of a collection of English conversations between users and three different conversational AI systems (Alana v2, Eliza, and CarbonBot), each with distinct goals. Each example is labeled by a minimum of three and a maximum of eight annotators\footnote{Following~\cite{vitsakis2023ilab}, we use binary format labels.}.

\subsection{LLMs}

We prompt four open-source instruction-tuned LLMs, each corresponding to a different model $M$: Olmo-7b-Instruct\footnote{\url{https://huggingface.co/allenai/OLMo-7B-Instruct}}, Llama-3-8b-Instruct\footnote{\url{https://huggingface.co/meta-llama/Meta-Llama-3-8B-Instruct}}, Gemma-7b-it\footnote{\url{https://huggingface.co/google/gemma-7b-it}} and Deepseek-7b-chat\footnote{\url{https://huggingface.co/deepseek-ai/deepseek-llm-7b-chat}}, applying methodology described in Section~\ref{sec:ICL}.
The original chat template is used for all models, along with a greedy search configuration, where~\emph{$do\_sample=False$}.

\subsection{Evaluation metrics} 
\label{subsec:eval_metrics}
This new multi-perspective paradigm, which embraces disagreements, calls for soft evaluation metrics\footnote{Soft metrics evaluate models' ability to predict both preferred and alternative interpretations, based on human judgments~\cite{rizzi-etal-2024-soft}. }
like Jensen-Shannon Divergence (JSD) and Cross-Entropy (CE)~\cite{peterson2019human, uma2021learning}. 
As such, we do not prioritize hard metrics such as precision and F1 scores, as they only focus only on the class with the highest probability and fail to capture nuance in~\emph{ambiguous instances} that are unclear or difficult to classify into a single class~\cite{wang2022capture}.

We use JSD as our main soft evaluation metric, which measures the distance between probability distributions,
alongside CE, which captures the model's confidence in its top prediction. This enables a comparative evaluation with the LeWiDi competition results~\cite{leonardelli2023semeval}.
Note that for soft evaluation, we convert aggregated and disaggregated hard labels to their soft counterparts only during the evaluation phase (Section~\ref{sec:label_space}).

We also report macro F1 scores to assess performance
across all classes equally, regardless of class imbalance, following the evaluation methods in the HS-Brexit and Conv-Abuse dataset papers~\cite{akhtar2021whose, cercas-curry-etal-2021-convabuse}. For the MD-Agreement dataset~\cite{leonardelli-etal-2021-agreeing}, the specific F1 score used is not specified.
Higher values indicate better performance for the hard metrics (accuracy and F1 score), while lower values are preferable for the soft metrics (JSD and CE).

\section{Results}

We assess multi-perspective ICL on various subjective tasks.  
We compare the baseline with the multi-perspective approach in both zero-shot and few-shot settings. In the zero-shot, we analyze the effect of the label space $l$ (Section~\ref{sec:label_space}), while in the few-shot setting, we examine the influence of demonstration examples $D$ (one-shot per class) through different selection and ordering strategies (Section~\ref{sec:demonstration_examples}). This section reports only the best-performing models, more detailed results are in Appendix~\ref{sec:zero_shot_results_overview} and~\ref{sec:few_shot_results}.

\begin{table}[!h]
\centering
\normalsize
\resizebox{0.95\columnwidth}{!}{ 
\begin{tabular}{lllcccc}
    \toprule
    \textbf{Dataset} & \textbf{LLM} & \textbf{Approach} & \textbf{Acc↑} & \textbf{F1↑} & \textbf{JSD↓} & \textbf{CE↓} \\
    \midrule

    \multirow{9}{*}{HS-Brexit} & \multirow{8}{*}{Deepseek-7b-chat} 
    & Baseline\_aggr\_0S        & 89.28 & 47.16 & 0.36 & 0.66 \\
    &  & Baseline\_aggr\_0S\_RL     & 88.09 & 46.83 & 0.26 & 0.46 \\
    &  & MultiP\_aggr\_0S          & 89.28 & \underline{64.93} & \textbf{0.19} & \underline{0.35} \\
    &  & MultiP\_aggr\_0S\_RL      & 86.90 & 50.64 & 0.28 & 0.50 \\
    &  & Baseline\_aggr\_FS        & 89.28 & \underline{52.15} & 0.21 & \underline{0.39} \\
    &  & Baseline\_aggr\_FS\_RL     & 86.90 & 46.49 & 0.21 & 0.43 \\
    &  & MultiP\_aggr\_FS         & 88.69 & \underline{51.74} & \textbf{0.19} & 0.42 \\
    &  & MultiP\_aggr\_FS\_RL     & 86.31 & 50.30 & 0.24 & 0.42 \\
    \midrule

    \multirow{9}{*}{MD-Agr} & \multirow{8}{*}{Deepseek-7b-chat} 
    & Baseline\_aggr\_0S        & 49.72 & 49.22 & 0.28 & 0.45 \\
    &  & Baseline\_aggr\_0S\_RL     & 45.14 & 43.42 & 0.28 & 0.47 \\
    &  & MultiP\_aggr\_0S          & 51.08 & 47.58 & 0.26 & 0.54 \\
    &  & MultiP\_aggr\_0S\_RL      & 66.69 & \underline{60.01} & \textbf{0.14} & \underline{0.43} \\
    &  & Baseline\_aggr\_FS        & 54.72 & 49.47 & 0.24 & 0.34 \\
    &  & Baseline\_aggr\_FS\_RL     & 57.11 & \underline{55.42} & 0.23 & 0.37 \\
    &  & MultiP\_aggr\_FS         & 51.78 & 47.35 & 0.25 & 0.34 \\
    &  & MultiP\_aggr\_FS\_RL     & 54.69 & 52.01 & \textbf{0.18} & \underline{0.25} \\
    \midrule

    \multirow{9}{*}{ConvAbuse} & \multirow{4}{*}{Deepseek-7b-chat} 
    & Baseline\_aggr\_0S        & 42.79 & 45.68 & 0.25 & 0.41 \\
    &  & Baseline\_aggr\_0S\_RL     & 52.71 & \underline{51.95} & \textbf{0.14} & \underline{0.29} \\
    &  & MultiP\_aggr\_0S         & 46.83 & 45.83 & 0.24 & 0.38 \\
    &  & MultiP\_aggr\_0S\_RL     & 53.14 & 45.09 & 0.18 & 0.32 \\
    \cmidrule{2-7}
    & \multirow{4}{*}{Olmo-7b-Instruct} 
    & Baseline\_aggr\_FS        & 46.73 & \underline{45.68} & 0.25 & 0.41 \\
    &  & Baseline\_aggr\_FS\_RL     & 50.73 & \underline{44.95} & \textbf{0.14} & \underline{0.29} \\
    &  & MultiP\_aggr\_FS         & 46.83 & \underline{45.83} & 0.24 & 0.38 \\
    &  & MultiP\_aggr\_FS\_RL     & 53.14 & \underline{45.09} & 0.18 & 0.32 \\

    \bottomrule
\end{tabular}
}
\vspace{-0.7em}
\caption{Zero-shot ($0S$) and Few-shot ($FS$) results for the best-performing LLMs. Few-shot uses BM-25 retrieval. $RL$ = role-playing, $aggr$ = aggregated labels. Best JSD scores in \textbf{bold}, best CE and F1 scores are \underline{underlined}.}

\label{tab:merged_results}
\vspace{-1em}
\end{table}

\subsection{Zero-shot}
\label{sec:results_zero_shot_appendix}
\paragraph{Impact of Label Space - Aggregated Labels}

~\emph{Results show that the multi-perspective approach outperforms the baseline in most zero-shot scenarios, proving to be the most effective prompt strategy, with lower JSD, lower CE and higher F1 scores} (Table \ref{tab:merged_results} and Table \ref{tab:results_zero_icl} in Appendix \ref{sec:zero_shot_results_overview}). This indicates that explicitly prompting LLMs with multiple perspectives and contextual instructions can be effective.

Our results show that DeepSeek-7b-chat outperforms Olmo-7b-Instruct, which is claimed to be fairer~\cite{groeneveld2024olmo}, across all three subjective tasks, demonstrating better classification and generalization performance (Table \ref{tab:results_zero_icl} in Appendix \ref{sec:zero_shot_results_overview}). Specifically, DeepSeek-7b-chat achieves the best JSD score (0.14) on MD-Agreement using the multi-perspective approach with role-playing for predicting aggregated hard labels. It matches this score on ConvAbuse using the baseline approach with role-playing. 
For HS-Brexit, the multi-perspective approach again performs best, achieving the highest JSD score (0.19). Overall, we observe that lower JSD generally aligns with lower CE. 
Further analysis reveals that DeepSeek-7b-chat is more prone to bias in hate speech detection than in abusive or offensive language detection.

\paragraph{Impact of Label Space - Disaggregated Labels}

\emph{Overall, we observe that LLMs perform poorly when predicting disaggregated labels, both hard and soft, struggling to capture the nuances of human disagreement. 
This leads to higher JSD and CE scores compared to aggregated label predictions, regardless of whether the baseline or multi-perspective approach is used to obtain aggregated labels} (Table~\ref{tab:results_zero_icl} in Appendix~\ref{sec:zero_shot_results_overview}).

These findings underscore ICL's limitations, particularly its reliance on the predefined label space~\cite{min2022rethinking}.
Specifically, when we explicitly prompt LLMs to predict soft labels, they tend to exhibit low variability, often favoring patterns like [0: 0.9, 1: 0.1] and [0: 0.8, 1: 0.2] as noted by~\citet{pavlovic-poesio-2024-effectiveness} in their study of GPT-3. 
A similar trend is observed in disaggregated hard label predictions, where LLMs tend to output monolithic discrete distributions, such as $[1, 0, 0, 0]$, $[0, 0, 0, 0]$, or $[1, 1, 1, 1]$. 
This shows LLMs struggle to capture the nuances of human disagreement with disaggregated hard and soft labels.

\subsection{Few-shot}
\label{sec:results_few_shot_appendix}
\paragraph{Impact of Demonstration Selection}
\label{sec:impact_select}
\emph{Results highlight that
two-stage ranking improves JSD and CE for both baseline and multi-perspective approaches. Specifically, for the multi-perspective approach, two-stage ranking performs better with disaggregated labels in both JSD and CE,  while showing comparable results for aggregated labels} (Table \ref{tab:results_bm25_re_rank}, \ref{tab:results_bm25_hs_md} and \ref{tab:results_bm25_conv} in Appendix \ref{sec:few_shot_results}).
We retrieve the demonstration examples
using three approaches with random ordering: text similarity, annotator disagreement and a two-stage ranking combining both.

Among the prompt strategies, the multi-perspective approach, when predicting aggregated labels, slightly outperforms the baseline in JSD and CE scores across all datasets, using both BM-25 and PLMs. 
However, BM-25 proves more effective for selecting similar examples in both the baseline and multi-perspective approaches, thus we use BM-25 for the following results.
Table~\ref{tab:merged_results} shows the best-performing models for each benchmark dataset. DeepSeek-7b-chat performs best on HS-Brexit and MD-Agreement, surpassing Olmo-7b-Instruct, Llama3-8b-Instruct, and Gemma-7b-it. The most effective strategy is the multi-perspective approach with aggregated labels, yielding low JSD scores of 0.19 and 0.18. However, for ConvAbuse, Olmo-7b-Instruct surpasses DeepSeek-7b-chat in JSD using the baseline approach, despite lower F1 scores (Tables~\ref{tab:results_bm25_hs_md} \& \ref{tab:results_bm25_conv} in Appendix~\ref{sec:few_shot_results}).

Selecting demonstration examples by annotator disagreement yields inconsistent results\footnote{Detailed results will be shared upon acceptance due to space constraints.}. We hypothesize that high-entropy examples introduce too much diversity or ambiguity, hindering the model's performance.
Moreover, approaches that leverage disaggregated labels consistently underperform compared to those using aggregated labels (both baseline and multi-perspective), even with careful example selection.

\begin{table}[!h]
\centering
\normalsize
\resizebox{0.98\columnwidth}{!}{ 
\begin{tabular}{lllcccc}
    \toprule
    \textbf{Dataset} & \textbf{LLM} & \textbf{Approach} & \textbf{Acc↑} & \textbf{F1↑} & \textbf{JSD↓} & \textbf{CE↓} \\
    \midrule
    \multirow{5}{*}{HS-Brexit} & \multirow{4}{*}{Deepseek.7b-chat}      
    & Baseline\_aggr\_FS        & 88.69 & 47.00 & 0.21 & \underline{0.39} \\
    &  & Baseline\_aggr\_RL & 87.50 & 46.67 & 0.22 & 0.42 \\
    &  & MultiP\_aggr\_FS      & 88.69 & 47.00 & \textbf{0.18} & 0.42 \\
    &  & MultiP\_aggr\_FS\_RL  & 87.50 & \underline{50.99} & 0.25 & 0.44   \\
 
    \midrule
    \multirow{5}{*}{MD-Agr} & \multirow{4}{*}{Deepseek.7b-chat}   
     & Baseline\_aggr\_FS\_CL        & 55.02 & 49.72 & 0.23 & 0.33   \\
    &  & Baseline\_aggr\_FS\_RL\_CL      & 57.98 & \underline{56.23} & 0.21 & 0.36  \\
    &  & MultiP\_aggr\_FS\_CL      & 51.78 & 47.27 & 0.25 & 0.33 
    \\
    &  & MultiP\_aggr\_FS\_RL\_CL  & 55.11 & 52.83 & \textbf{0.17} & \underline{0.24} \\

    \midrule
    \multirow{5}{*}{ConvAbuse} & \multirow{4}{*}{Olmo-7b-Instruct}      
      & Baseline\_aggr\_FS\_CL         & 75.95 & 53.27 & \textbf{0.15} & \underline{0.25}  \\
    &                                       & Baseline\_aggr\_FS\_RL\_CL      & 71.62 & 47.14 & 0.20 & 0.33  \\
    &                                       & MultiP\_aggr\_FS\_CL      & 71.17 & 46.01 & \textbf{0.15} & 0.30  \\
    &                                       & MultiP\_aggr\_FS\_RL\_CL      & 58.33 & \underline{55.44} & \textbf{0.14} & 0.44  \\
 
    \bottomrule
\end{tabular}
}
\vspace{-0.6em}
\caption{Few-shot ($FS$) results using BM-25 to retrieve the most similar examples, ordered with curriculum learning ($CL$), for the best-performing model (DeepSeek-7b-chat). $RL$ denotes role-playing and $aggr$ aggregated labels. Best JSD scores are 
in \textbf{bold}, best CE and F1 scores are \underline{underlined}.
}
\label{tab:results_few_bm25_cl}
\vspace{-1.0em}
\end{table}

\begin{table}[!t]

\centering
\small
\resizebox{0.89\columnwidth}{!}{ 
\begin{tabular}{llcc}
    \toprule
    \textbf{Dataset} & \textbf{Approach} & \textbf{\textit{micro} F1↑} & \textbf{CE↓} \\
    \midrule
    \multirow{8}{*}{HS-Brexit}     
    & SE (Baseline)        & 84.23 & 2.72 \\
    &  SE (Best) & \textbf{92.86} & \textbf{0.24} \\
    & MultiP\_aggr\_$0S$ (Best) &  \underline{89.29} & \underline{0.35} \\
      & Baseline\_aggr\_$FS$ (Best) & \underline{89.28} & 0.39 \\
     \cline{2-4} 
    &  GPT-3.5\_disaggr\_soft\_$0S$ & 69.60 & 5.04 \\
    & MultiP\_disaggr\_soft\_$0S$ (Best) & \underline{70.24}  & \textbf{0.52} \\
    & MultiP\_disaggr\_soft\_$FS$ (Best) & \textbf{83.33} & \underline{0.97} \\
    \hline \hline 
    \multirow{8}{*}{MD-Agr}
     & SE (Baseline)      & 53.38 & 7.39 \\
    &  SE (Best) & \textbf{84.61} & 0.47 \\
    & Baseline\_aggr\_$0S$ (Best) & 72.53 & \underline{0.35} \\
    & MultiP\_aggr\_$FS$ (Best)   & \underline{75.23} & \textbf{0.24} \\
    \cline{2-4}
    & GPT-3.5\_disaggr\_soft\_$0S$ & \underline{52.00} & 3.83 \\
    & MultiP\_disaggr\_soft\_$0S$ (Best)   & \textbf{74.81} & \textbf{0.32} \\
    & MultiP\_disaggr\_soft\_$FS$ (Best)   & 49.91 & \underline{1.25} \\
    \hline \hline
    \multirow{8}{*}{ConvAbuse} 
    & SE (Baseline)        & 74.09 & 3.48 \\
    & SE (Best) & \textbf{94.16} & \textbf{0.19} \\
     & Baseline\_aggr\_$0S$ (Best)   & \underline{82.02} & \underline{0.23} \\
    & MultiP\_aggr\_$FS$ (Best)   & 71.52 & \underline{0.23} \\
    \cline{2-4}
    &  GPT-3.5\_disaggr\_soft\_$0S$ & \textbf{90.20} & 3.75 \\
    & MultiP\_disaggr\_soft\_$0S$ (Best) & 73.46 & \underline{0.42}\\
    & MultiP\_disaggr\_soft\_$FS$ (Best) & \underline{77.21} &  \textbf{0.39}\\ 
    \bottomrule
\end{tabular}
}
\vspace{-0.8em}
\caption{The comparison of our best $aggr$ and $disaggr$ $0S$ and $FS$ models: with the SE fine-tuned baseline and best models, followed by GPT-3.5-turbo ($0S$). For the SE and GPT-3.5-turbo, the best is highlighted in \textbf{bold}, and the second-best is \underline{underlined} separately.}
\label{tab:performance_lewidi}
\vspace{-1.0em}
\end{table}

\paragraph{Impact of Demonstration Ordering}
 \emph{Our findings reveal that the ordering has a limited impact on model performance} (Table \ref{tab:results_few_bm25_cl}).
We assess two strategies: random ordering and curriculum learning (CL) (Section \ref{sec:order}).
Examples are randomly ordered across all three demonstration selection approaches, while CL is applied only to BM-25, the most effective selection approach (see Section \ref{sec:impact_select}).
While CL shows slight improvements over random ordering for disaggregated hard and soft label predictions, these gains are inconsistent and vary across datasets, suggesting limited generalizability.

\paragraph{Overall Findings}

Our results are comparable to the SE LeWiDi competition~\cite{leonardelli2023semeval}, which uses fine-tuned BERT-based models, and to~\citet{pavlovic-poesio-2024-effectiveness}, which employ GPT-3.5-turbo $0S$ for predicting disaggregated labels (Table~\ref{tab:performance_lewidi}). Both the baseline model and LeWiDi’s best model predict aggregated labels (majority), with the latter using an ensemble trained on labels from different annotators.  
For a fair comparison, we contrast our best $0S$ and $FS$ aggregated models with LeWiDi and disaggregated soft models with GPT-3.5.

Our primary evaluation is based on JSD as the soft metric (along with CE), plus macro F1 as the hard metric. However, in Table~\ref{tab:performance_lewidi}, we report CE scores alongside micro F1.
This enables a comparison of our best models with state-of-the-art results, as these studies use micro F1. Micro F1 assesses overall model performance, but may miss class imbalance. Micro F1 scores are relative, as we select the best models by lowest CE. 
The results reveal that model performance is heavily dependent on the dataset. We compare our results with LeWiDi and GPT-3.5 separately for each dataset. For HS-Brexit, our best $0S$ model ranks second in CE, in MD-Agreement we obtain both the best and the second-best CE scores, outperforming LeWiDi's best models. 
For ConvAbuse, both our $0S$ and $FS$ models achieve the second-best CE scores.
In terms of micro F1, we rank second for all three datasets compared to LeWiDi's best models.
Our best disaggregated models outperform GPT-3.5 in CE across all three datasets. In terms of micro F1, we obtain the best performance for HS-Brexit and MD-Agreement, while ranking second for ConvAbuse.

\section{Conclusion \& Future Work}
In this work, we evaluated how well LLMs capture multiple perspectives and handle annotator disagreement in subjective NLP tasks. Our research highlights the critical role of social intelligence in aligning LLMs with diverse user preferences. By comparing standard prompting with different multi-perspective ICL approaches, we found that explicitly guiding models to consider diverse viewpoints improves model performance in both zero-shot and few-shot settings. Overall, findings suggest that demonstration selection and ordering do not always enhance subjective task performance. Future work will expand model evaluations by integrating bias evaluation metrics alongside traditional performance measures.

\clearpage
\section{Limitations}

This study is subject to several limitations. First, our analysis is constrained by the limited availability of resources, particularly given that perspectivism is an emerging research paradigm. Consequently, our evaluation relies on datasets from the LeWiDi competition, which currently represent the most prominent benchmarks within perspectivist NLP.
Our focus on subjective tasks—hate speech, offensive language, and abusive language detection—was driven largely by the availability of suitable datasets. These datasets are typically restricted to binary classification which hinders the generalization of our approach to more complex, multi-class scenarios.
Moreover, most datasets are monolingual, primarily in English, which limits the applicability of our findings in multilingual settings. We plan to explore data collection and annotation strategies grounded in the perspectivist paradigm, with the main goal of developing benchmarks for low-resource languages.

Future work will aim to extend this methodology to a broader spectrum of model architectures and access modalities, including both open-source and proprietary LLMs such as Claude\footnote{\url{https://docs.anthropic.com/it/docs/welcome}} and GPT-based models\footnote{\url{https://openai.com/index/gpt-4/}}.

\paragraph{Ethics Statement}

Modeling human subjectivity is inherently connected to social bias since annotators’ backgrounds, experiences, and personal values shape their annotations, influencing both model training and the evaluation of LLMs. Recognizing the societal impact of these technologies is essential, as they risk amplifying dominant viewpoints, potentially disadvantaging marginalized groups. Incorporating minority perspectives is crucial to ensuring diverse voices are acknowledged and not overshadowed in a landscape often shaped by prevailing opinions.

\paragraph{Impact Statement}

Our research is driven by the need to mitigate societal biases that may inadvertently be embedded in NLP systems and broader AI technologies. By focusing on LLMs, we aim to promote inclusivity and diversity through multi-perspective prompting. This study highlights the challenges of applying LLMs to subjective tasks and encourages the NLP community to develop perspective-aware models that are more inclusive, responsible, and socially intelligent to be able for effective human-machine interaction.

\bibliography{custom,anthology}
\clearpage
\begin{sidewaystable}[!t]
    \centering
    \hspace{-2.3em}
    \tiny
    \rowcolors{2}{white}{blue!15} 
    \begin{tabular}{p{5cm} c c c c c c}  
        \toprule
        \textbf{Input} & \textbf{Original\_Labels} & \textbf{Topic} & \textbf{Baseline} & \textbf{MP\_aggr\_hard} & \textbf{MP\_disaggr\_hard} & \textbf{MP\_disaggr\_soft} \\
        \midrule
        \rowcolor{blue!15} Expel the cop unions & [0,1,1,0,0] & \#blm & 0 & 1 & [0,0,0,0,0] & [0.0,1.0] \\
        \rowcolor{blue!15} Right-wing Accelerationists and Infiltrators = White Supremacists at Protests. & [1,0,0,0,1] & \#blm & 1 & 1 & [1,0,0,0,0] & [1.0,0.0] \\
        \rowcolor{blue!15} Why does the Government Far Left, only pick on the Blacks, when there are so many other races in America? & [0,1,0,1,0] & \#blm & 0 & 0 & [0,0,0,0,0] & [1.0,0.0] \\
        \rowcolor{green!15} And this Big Guy will mindlessly drug hoover all the Cocaine off Malia Obama in the West Wing kitchen before doing megadeals on China White. & [1,0,1,0,0]  & \#election2020 & 1 & 1 & [1,0,0,0,0] & [0.0,1.0] \\
        \rowcolor{green!15} So you admit that Trump was also classless? I'd say that failing to condemn white supremacists is pretty classless. Yet you pretended not to understand why he failed to do it. Brian, just admit it - the president is also a white nationalist. & [0,1,0,0,1]  & \#election2020 & 1 & 1 & [1,0,0,0,1] & [0.8,0.2] \\
        \rowcolor{green!15} Hey bruh, that's not cool. Democrats play dirty but we should play smarter. \#MAGA & [0,0,1,1,0]  & \#election2020 & 1 & 0 & [1,0,0,0,0] & [1.0,0.0] \\
        \rowcolor{pink!15} This is what happens in the brains of people who believe that 5g internet causes Covid-19. & [1,0,0,1,0]  & \#covid19 & 0 & 1 & [1,0,0,0,0] & [1.0,0.0] \\
        \rowcolor{pink!15} Covid 19-poll? You're irrelevant now. Shut up & [1,1,0,0,0] & \#covid19 & 0 & 1 & [1,0,1,0,0] & [1.0,0.0] \\
        \rowcolor{pink!15} This is disgusting! Just like his dad! & [1,0,0,0,1] & \#covid19 & 0 & 0 & [0,0,0,0,0] & [1.0,0.0] \\
        \bottomrule
    \end{tabular}
    \vspace{-0.5em}
    \caption{Examples from the MD-Agreement dataset with baseline and multi-perspective predictions, along with topic-based coloring. MP stands for Multi-Perspective, \textit{aggr.} for aggregated, and \textit{disaggr.} for disaggregated.}
    \label{tab:md_multip_labels_colored}
    \vspace{-1.0em}
\end{sidewaystable}

\section*{Appendix}
\label{sec:appendix}
\setcounter{subsection}{0} 
\renewcommand{\thesubsection}{\Alph{subsection}}

\subsection{Error analysis}

We conduct a qualitative analysis comparing the baseline and multi-perspective approaches, focusing on both aggregated and disaggregated labels. Our analysis is centered on the MD-Agreement dataset, which has the highest level of annotator disagreement, with 42\% of instances showing full agreement (Table~\ref{tab:datasets}). Specifically, we examine instances of~\emph{weak agreement}~\cite{leonardelli-etal-2021-agreeing}, where at least two out of five annotators disagree.

The MD-Agreement dataset contains 3057 instances, with disagreements between the multi-perspective and baseline predictions found in 973 instances. Table~\ref{tab:md_multip_labels_colored} presents examples of weak agreement for each topic, detailing input, individual label distribution, topic, and predictions. Mismatches, primarily false positives, occur when the baseline correctly predicts the majority label (not-offensive) while the multi-perspective approach aligns with the minority label, indicating closer alignment with human label distribution and reducing JSD. These false positives often involve implicit offensive content, like irony. However, when slur words are explicit, the model’s predictions become more consistent, with only 5.30\% mismatches.
Interestingly, multi-perspective approaches that predict disaggregated labels—whether as individual annotations (hard) or probability scores (soft)—struggle to accurately capture human disagreement.
As shown in Table~\ref{tab:md_multip_labels_colored}, disaggregated predictions (both hard and soft) exhibit minimal variation. For instance, in the first data point, when the original hard labels are $[0,1,1,0,0]$, the model predicts $[0,0,0,0,0]$, overlooking the disagreements. Similarly, the soft label predictions often favor bimodal outputs\footnote{The model is highly confident in two different possible outcomes.}. For example, in the first case, the multi-perspective model predicts $[0.0,1.0]$ as the probability distribution.

These findings highlight the limitations of LLMs in capturing human disagreement within disaggregated labels. While LLMs are effective in predicting aggregated labels, they struggle to represent the nuanced disagreements present in~\emph{ambiguous instances}.
This underscores the need for further advancements in developing perspective-aware LLMs that can more effectively model and learn from human-like disagreement.

\clearpage
\subsection{Zero-shot results}
\label{sec:zero_shot_results_overview} 
In this section, we present the results of applying our approach in a zero-shot setting. A detailed discussion of the impact of label space- aggregated and disaggregated labels- can be found in Section \ref{sec:results_zero_shot_appendix}.
\renewcommand{\arraystretch}{0.8} 
\setlength{\tabcolsep}{3pt} 
\begin{table}[H]
\centering
\tiny
\begin{tabular}{lllcccc}
    \toprule
    \textbf{Dataset} & \textbf{LLM} & \textbf{Approach} & \textbf{Acc↑} & \textbf{F1↑} & \textbf{JSD↓} & \textbf{CE↓} \\
        \midrule
    \multirow{18}{*}{HS-Brexit} 
    & \multirow{3}{*}{Olmo-7b-Instruct}      & Baseline\_aggr\_0S         & 55.95 & 49.01 & 0.40 & 1.21 \\
    &                                       & Baseline\_aggr\_0S\_RL      
        & 57.26 & 50.75 & 0.41 & 1.23 \\
    &                                       & MultiP\_aggr\_hard 0S      & 40.47 & 37.61 & 0.43 & 1.18 \\
    &                                       & MultiP\_aggr\_hard  0S\_RL      &  64.28 & \underline{53.81} & \textbf{0.38} & \underline{1.08} \\
     &                                       & MultiP\_disaggr\_soft 0S      & 25.59 & 24.99 & 0.59 & 11.67 \\
       &                                       & MultiP\_disaggr\_soft 0S\_RL      & 29.16 & 27.81 & 0.54 & 7.68  \\
     &                                       & MultiP\_disaggr\_hard 0S      & 28.57 & 27.32 & 0.59 & 14.01  \\
      &                                       & MultiP\_disaggr\_hard 0S\_RL      & 32.14 & 30.72 & 0.57 & 13.61  \\
    \cmidrule(lr){2-7}
    & \multirow{3}{*}{Gemma-7b-it}          & Baseline\_aggr\_0S         & 75.59 & 51.02 & \textbf{0.26} & \underline{0.48} \\
    &                                       & Baseline\_aggr\_0S\_RL      & 74.34 & 52.63 & 0.28 & 0.51 \\
    &                                       & MultiP\_aggr\_hard 0S      &  67.85 & 53.21 & 0.30 & 0.67  \\
    &                                       & MultiP\_aggr\_hard 0S\_RL      & 23.21 & 23.18 & 0.49 & 1.26 \\
     &                                       & MultiP\_disaggr\_soft 0S      & 70.23 & \underline{57.48} & 0.28 & 0.52 \\
       &                                       & MultiP\_disaggr\_soft 0S\_RL      & 61.31 & 53.45 & 0.31 & 0.57  \\
     &                                       & MultiP\_disaggr\_hard 0S      & 26.78 & 25.12 & 0.61 & 15.01  \\
      &                                       & MultiP\_disaggr\_hard 0S\_RL      & 13.09 & 12.89 & 0.71 & 17.51  \\
    \cmidrule(lr){2-7}
    & \multirow{3}{*}{Llama3-8b-Instruct}   & Baseline\_aggr\_0S         & 68.45 & 59.49 & 0.28 & 0.91 \\
    &                                       & Baseline\_aggr\_0S\_RL      & 65.47 & 57.32 & 0.29 & 0.97 \\
    &                                       & MultiP\_aggr\_hard 0S      & 60.11 & 53.09 & 0.33 & 0.99  \\
    &                                       & MultiP\_aggr\_hard 0S\_RL      & 76.78 & \underline{66.53} & \textbf{0.23} & \underline{0.65} \\
     &                                       & MultiP\_disaggr\_soft 0S      & 61.90 & 54.44 & 0.27 & 0.93 \\
       &                                       &MultiP\_disaggr\_soft 0S\_RL      & 58.92 & 52.21 & 0.28 & 1.92  \\
     &                                       & MultiP\_disaggr\_hard 0S      & 13.09 & 12.48 & 0.71 & 17.51  \\
      &                                       & MultiP\_disaggr\_hard 0S\_RL      & 15.47 & 15.42 & 0.68 & 16.98  \\
    \midrule
    \multirow{18}{*}{MD-Agr} 
    & \multirow{3}{*}{Olmo-7b-Instruct}      & Baseline\_aggr\_0S         & 64.57 & 64.24 & 0.47 & 0.60 \\
    &                                       & Baseline\_aggr\_0S\_RL      & 66.40 & 65.39 & 0.42 & 0.63  \\
    &                                       & MultiP\_aggr\_hard 0S      & 67.28 & 66.04 & \textbf{0.31} & \underline{0.47} \\
     &                                       & MultiP\_aggr\_hard 0S\_RL      & 69.77 & \underline{67.89} & 0.41 & 0.57 \\
       &                                       & MultiP\_disaggr\_soft 0S      & 74.28 & 61.70 & \textbf{0.32} & 1.45 \\
      &                                       & MultiP\_disaggr\_soft 0S\_RL      & 76.31 & 63.78 & 0.35 & 1.49  \\
       &                                       & MultiP\_disaggr\_hard 0S      & 54.23 & 53.80 & 0.64 &  2.20 \\
        &                                       & MultiP\_disaggr\_hard 0S\_RL      & 55.33 & 56.03 & 0.58 & 2.16 \\  
    \cmidrule(lr){2-7}
    & \multirow{3}{*}{Gemma-7b-it}          & Baseline\_aggr\_0S         & 52.43 & 51.64 & 0.52 & 0.58 \\
    &                                       & Baseline\_aggr\_0S\_RL      & 64.01 & \underline{62.47} & \textbf{0.26} & 0.35 \\
    &                                       & MultiP\_aggr\_hard 0S      & 58.35 & 58.35 & \textbf{0.25} & 0.39  \\
    &                                       & MultiP\_aggr\_hard 0S\_RL      & 55.18 & 55.08 & \textbf{0.26} & 0.36 \\
     &                                       & MultiP\_disaggr\_soft 0S     & 62.08 & \underline{62.02} & \textbf{0.26} & \underline{0.32} \\
     &                                       & MultiP\_disaggr\_soft 0S\_RL     & 61.78 & 60.66 & 0.27 & 0.40\\
      &                                       & MultiP\_disaggr\_hard 0S     & 57.63 & 57.48 & 0.50 & 1.33\\
       &                                       & MultiP\_disaggr\_hard 0S\_RL     & 54.66 & 45.64 & 0.66 & 1.37\\
    \cmidrule(lr){2-7}
    & \multirow{3}{*}{Llama3-8b-Instruct}   & Baseline\_aggr\_0S        & 62.70 & 62.45 & \textbf{0.16} & 0.52  \\
    &                                       & Baseline\_aggr\_0S\_RL      & 68.69 & \underline{66.47} & 0.31 & \underline{0.44} \\
    &                                       & MultiP\_aggr\_hard 0S      & 51.88 & 51.06 & 0.45 & 0.58 \\
     &                                       & MultiP\_aggr\_hard 0S\_RL      & 70.85 & 63.52 & 0.32 & 0.47  \\
       &                                       & MultiP\_disaggr\_soft 0S     & 71.54 & 63.26 & 0.41 & 1.39 \\
        &                                       & MultiP\_disaggr\_soft 0S\_RL     & -- & -- & -- & -- \\
        &                                       & MultiP\_disaggr\_hard 0S     & 50.85 & 48.69 & 0.60 &  2.25 \\
          &                                       & MultiP\_disaggr\_hard 0S\_RL     & 40.66 & 39.89 & 0.64 & 2.31 \\
    \midrule
    \multirow{18}{*}{ConvAbuse} 
    & \multirow{3}{*}{Olmo-7b-Instruct}      & Baseline\_aggr\_0S        & 78.69 & 63.00 & \textbf{0.21} & 0.32 \\
    &                                       & Baseline\_aggr\_0S\_RL      & 77.50 & \underline{64.93} & 0.42 & 0.53 \\
    &                                       & MultiP\_aggr\_hard 0S      & 75.23 & 61.81 & 0.42 & 0.54  \\
    &                                       & MultiP\_aggr\_hard 0S\_RL      & 76.66 & 62.23 & 0.31 & \underline{0.41} \\
     &                                       & MultiP\_disaggr\_soft 0S      & 83.52 & 52.12 & 0.42 & 1.57 \\
       &                                       & MultiP\_disaggr\_soft 0S\_RL      & 85.12 & 55.35 & 0.41 & 1.43  \\
     &                                       & MultiP\_disaggr\_hard 0S      & 59.28 & 54.83 & 0.36 & 2.50  \\
      &                                       & MultiP\_disaggr\_hard 0S      & 53.09 & 44.25 & 0.49 & 3.13  \\
    \cmidrule(lr){2-7}
    & \multirow{3}{*}{Gemma-7b-it}          & Baseline\_aggr\_0S         & 74.16 & 57.18 & 0.65 & 0.77 \\
    &                                       & Baseline\_aggr\_0S\_RL      & 77.02 & \underline{62.29} & \textbf{0.16} & \underline{0.23} \\
    &                                       & MultiP\_aggr\_hard 0S      & 74.64 & 61.80 & 0.19 & 0.31 \\
    &                                       & MultiP\_aggr\_hard 0S\_RL      & 74.64 & \underline{62.27} & \textbf{0.15} & 0.28 \\
    &                                       & MultiP\_disaggr\_soft  0S      & 78.92 & 51.18 & 0.38 & 1.81 \\
    &                                       & MultiP\_disaggr\_soft 0S\_RL      & 79.54 & 59.32 & 0.35 & 1.24 \\
    &                                       & MultiP\_disaggr\_hard 0S      & 51.78 & 50.38 & 0.40 & 2.55 \\
     &                                       & MultiP\_disaggr\_hard 0S\_RL      & 52.13 & 41.88 & 0.37 & 2.38 \\
    \cmidrule(lr){2-7}
    & \multirow{3}{*}{Llama3-8b-Instruct}   & Baseline\_aggr\_0S         & 76.69 & 63.36 & 0.32 & 0.42 \\
    &                                       & Baseline\_aggr\_0S\_RL      & 67.15 & \underline{66.10} & 0.30 & 0.41  \\
    &                                       & MultiP\_aggr\_hard 0S      & 80.71 & 61.68 & 0.33 & 0.42 \\
     &                                       & MultiP\_aggr\_hard 0S\_RL      & 81.42 & 61.27 & \textbf{0.28} & \underline{0.32} \\
      &                                       & MultiP\_disaggr\_soft 0S      & 81.78 & 52.63 & 0.34 & 0.47 \\
       &                                       & MultiP\_disaggr\_soft 0S\_RL      & 83.97 & 63.33 & 0.32 & 0.42 \\
        &                                       & MultiP\_disaggr\_hard 0S      & 63.85 & \underline{65.69} & 0.42 & 3.10 \\
        &                                       & MultiP\_disaggr\_hard 0S\_RL      & 51.67 & 42.52 & 0.37 &  2.59 \\

    \bottomrule
\end{tabular}
\caption{Evaluation results of Zero-shot ($0S$) ICL for Olmo-7b-Instruct, Gemma-7b-it and Llama3-8b-Instruct on HS-Brexit, MD-Agreement, and ConvAbuse datasets. Llama3-8B-Instruct refuses to predict soft labels with $RL$ on the MD-agreement dataset, likely due to the task's high subjectivity and the uncertainty of probability-based labels. $RL$ denotes role-playing, $aggr$ aggregated and $disaggr$ disaggregated labels. The best JSD scores are highlighted in \textbf{bold}, and the best F1 and CE scores are \underline{underlined}.}
\label{tab:results_zero_icl}
\end{table} 
 
\vspace{0.5cm}
\clearpage

\subsection{Few-shot results}
\label{sec:few_shot_results}
In this section, we present the results of applying our approach in a few-shot setting. A detailed discussion of the impact of demonstration selection and ordering can be found in Section \ref{sec:results_few_shot_appendix}.
\renewcommand{\arraystretch}{0.8} 
\setlength{\tabcolsep}{3pt} 
\begin{table}[!ht]
\centering
\tiny
\begin{tabular}{@{}lllcccc@{}}
    \toprule
    \textbf{Dataset} & \textbf{LLM} & \textbf{Approach} & \textbf{Acc↑} & \textbf{F1↑} & \textbf{JSD↓} & \textbf{CE↓}\\
    \midrule
    \multirow{18}{*}{HS-Brexit} 
    & \multirow{3}{*}{Olmo-7b-Instruct} & Baseline\_aggr\_FS        & 57.14 & 48.14 & \textbf{0.34} & 0.87\\
    &  & Baseline\_aggr\_FS\_RL & 56.23 & 43.37 & 0.38 & 0.98   \\
    &  & MultiP\_aggr\_hard FS      &  55.95 & 47.32 & \textbf{0.34} & \underline{0.85} \\
     &  & MultiP\_aggr\_hard FS\_RL & 54.76 & 46.51 & 0.36 & 0.91      \\
       &  & MultiP\_disaggr\_soft FS  & 64.88 & 47.31 & 0.38 & 4.02     \\
      &  & MultiP\_disaggr\_soft FS\_RL  & 70.14 & \underline{51.74} & 0.37 & 3.35\\
       &                                       & MultiP\_disaggr\_hard FS      & 37.50 & 33.11 & 0.53 & 11.11 \\
        &                                       & MultiP\_disaggr\_hard FS\_RL  & 27.38 & 26.87 & 0.62 & 14.29 \\

    \cmidrule(lr){2-7}
    & \multirow{3}{*}{Gemma-7b-it}  & Baseline\_aggr\_FS & 83.93 & 52.02 & 0.30 & 0.55  \\
    &  & Baseline\_aggr\_FS\_RL &  82.49 & 51.78 & 0.36 & 0.66 \\
    &  & MultiP\_aggr\_hard FS      & 76.78 & 51.76 & \textbf{0.27} & \underline{0.51}   \\
     &   & MultiP\_aggr\_hard FS\_RL & 68.45 & 52.67 & 0.34 & 0.67     \\
       &  & MultiP\_disaggr\_soft FS  & 77.97 & 66.96 & 0.28 & 1.06  \\
      &  & MultiP\_disaggr\_soft FS\_RL & 83.33 & \underline{70.00} & \textbf{0.26} &  0.97        \\
       & & MultiP\_disaggr\_hard FS  &  17.26 & 17.18 & 0.70 & 16.77\\
        & & MultiP\_disaggr\_hard FS\_RL  & 18.45 & 18.44 & 0.69 & 16.34 \\
    \cmidrule(lr){2-7}
    & \multirow{3}{*}{Llama3-8b-Instruct}   & Baseline\_aggr\_FS         & 58.92 & \underline{51.69} & \textbf{0.33} & 1.46  \\
    &  & Baseline\_aggr\_FS\_RL & 56.77 & 49.28 & 0.39 & 1.83\\
    &  & MultiP\_aggr\_hard FS     &  55.35 & 49.07 & 0.36 & \underline{1.37}\\
    & & MultiP\_aggr\_hard FS\_RL      & 56.54 & 49.94 & 0.37 & 1.58\\
    & & MultiP\_disaggr\_soft FS      & 58.33 & \underline{51.25} & 0.36 &  3.52 \\
    & & MultiP\_disaggr\_soft FS\_RL & 55.35 & 49.07 & 0.39 & 4.19  \\
    &  & MultiP\_disaggr\_hard FS   &  55.35 & 49.07 & 0.40 & 8.38\\
    &  & MultiP\_disaggr\_hard FS\_RL & 22.61 & 22.61 & 0.64 & 15.51\\
\midrule
    \multirow{18}{*}{MD-Agr} 
    & \multirow{3}{*}{Olmo-7b-Instruct}      & Baseline\_aggr\_FS        & 67.12 & 61.45 & 0.45 & 0.62  \\
    &                                       & Baseline\_aggr\_FS\_RL      & 66.30 & 62.61 & 0.43 & 0.51  \\
    &                                       & MultiP\_aggr\_hard FS      & 66.30 & 60.05 & 0.41 & 0.54  \\
     &                                       & MultiP\_aggr\_hard FS\_RL      & 65.39 & 42.96 & \textbf{0.26} & \underline{0.38}  \\
       &                                       & MultiP\_disaggr\_soft FS      & 67.15 & \underline{64.61} & 0.43 & 1.25  \\
      &                                       & MultiP\_disaggr\_softFS\_RL      & 60.51 & 56.92 & 0.62 & 2.63  \\
       &                                       & MultiP\_disaggr\_hard FS      & 58.45 & 57.03 & 0.61 & 2.29 \\
        &                                       & MultiP\_disaggr\_hard FS\_RL      & 51.17 & 49.82 & 0.59 & 2.15  \\

    \cmidrule(lr){2-7}
    & \multirow{3}{*}{Gemma-7b-it}          & Baseline\_aggr\_FS        & 53.84  & 48.69 & \textbf{0.26}  & \underline{0.36}  \\
    &                                       & Baseline\_aggr\_FS\_RL      & 56.03 & \underline{55.78} & 0.28 & 0.43  \\
    &                                       & MultiP\_aggr\_hard FS      & 66.63 & 40.45 & \textbf{0.26} & 0.42  \\
    &                                       & MultiP\_aggr\_hard FS\_RL      & 61.95 & 48.42 & \textbf{0.26} & 0.41  \\
     &                                       & MultiP\_disaggr\_soft FS     & 58.66 & 53.71 & 0.50 & 2.27  \\
     &                                       & MultiP\_disaggr\_soft FS\_RL     & 59.27 & \underline{55.48} & 0.49 & 2.53  \\
      &                                       & MultiP\_disaggr\_hard FS     & 50.39 & 45.68 & 0.62 & 5.90  \\
       &                                       & MultiP\_disaggr\_hard FS\_RL     & 52.43 & 46.25 & 0.57 & 5.59  \\
    \cmidrule(lr){2-7}
    & \multirow{3}{*}{Llama3-8b-Instruct}   & Baseline\_aggr\_FS         & 50.76 & 49.96 & 0.28 & \underline{0.34}  \\
    &                                       & Baseline\_aggr\_FS\_RL      & 54.72 & 54.06 & 0.28 & 0.44  \\
    &                                       & MultiP\_aggr\_hard FS      & 59.24 & \underline{58.43} & 0.28 & 0.41  \\
     &                                       & MultiP\_aggr\_hard FS\_RL      & 59.37 & \underline{59.22} & \textbf{0.26} & \underline{0.35}  \\
       &                                       & MultiP\_disaggr\_soft FS     & 57.63 & 57.60 & 0.36 & 1.80 \\
        &                                       & MultiP\_disaggr\_soft FS\_RL     & 58.28 & 55.17 & 0.38 & 1.86  \\
        &                                       & MultiP\_disaggr\_hard FS     & 45.76 & 44.52 & 0.64 & 3.71  \\
          &                                       & MultiP\_disaggr\_hard FS\_RL     & 50.29 & 49.26 & 0.67 & 3.46  \\
    \midrule
    \multirow{18}{*}{ConvAbuse} 
    & \multirow{3}{*}{Olmo-7b-Instruct}      & Baseline\_aggr\_FS         & 74.04 & 65.61 & 0.43 & 0.56  \\
    &                                       & Baseline\_aggr\_FS\_RL      & 75.19 & \underline{66.61} & 0.38 & 0.44  \\
    &                                       & MultiP\_aggr\_hard FS      & 70.95 & 45.94 & \textbf{0.16} & \underline{0.31}  \\
    &                                       & MultiP\_aggr\_hard FS\_RL      & 64.16 & 58.03 & 0.32 & 0.42  \\
     &                                       & MultiP\_disaggr\_soft FS      & 57.89 & 51.89 & 0.18 & 0.45  \\
       &                                       & MultiP\_disaggr\_soft FS\_RL      & 58.51 & 53.37 & 0.41 & 0.85  \\
     &                                       & MultiP\_disaggr\_hard FS      & 49.78 & 43.24 & 0.60 & 2.93  \\
      &                                       & MultiP\_disaggr\_hard FS\_RL      & 49.34 & 43.93 & 0.58 & 2.89  \\
    \cmidrule(lr){2-7}
    & \multirow{3}{*}{Gemma-7b-it}          & Baseline\_aggr\_FS         & 78.33  & 51.65 & \textbf{0.15}  & \underline{0.29}  \\
    &                                       & Baseline\_aggr\_FS\_RL      & 81.42 & 49.47 & 0.20 & 0.32  \\
    &                                       & MultiP\_aggr\_hard FS      & 80.47 & 45.76 & \textbf{0.15} & \underline{0.28}  \\
    &                                       & MultiP\_aggr\_hard FS\_RL      & 57.14 & 53.98 & 0.20 & 0.34  \\
     &                                       & MultiP\_disaggr\_soft FS     & 64.57 & \underline{58.69} & 0.31 & 0.53  \\
     &                                       & MultiP\_disaggr\_soft FS\_RL     & 65.11 & \underline{59.35} & 0.27 & 0.39  \\
      &                                       & MultiP\_disaggr\_hard FS     & 50.31 & 49.84 & 0.66 & 4.19  \\
       &                                       & MultiP\_disaggr\_hard FS\_RL     & 52.71 & 50.02 & 0.61 & 3.92  \\
       
    \cmidrule(lr){2-7}
    & \multirow{3}{*}{Llama3-8b-Instruct}   & Baseline\_aggr\_FS         & 79.04 & 70.18 & 0.27 & 0.34  \\
    &                                       & Baseline\_aggr\_FS\_RL      & 70.59 & 65.85 & 0.22 & 0.36  \\
    &                                       & MultiP\_aggr\_hard FS      & 83.80 & \underline{77.29} & \textbf{0.14} & \underline{0.23}  \\
     &                                       & MultiP\_aggr\_hard FS\_RL      & 72.14 & 66.83 & 0.21 & 0.35  \\
      &                                       & MultiP\_disaggr\_soft FS      & 56.10 & 52.29 & 0.41 & 2.21  \\
       &                                       & MultiP\_disaggr\_soft FS\_RL      & 57.53 & 57.41 & 0.39 & 1.46  \\
        &                                       & MultiP\_disaggr\_hard FS      & 52.78 & 44.91 & 0.43 & 4.89  \\
        &                                       & MultiP\_disaggr\_hard FS\_RL      & 52.12 & 48.24 & 0.41 & 4.79  \\
    \bottomrule
\end{tabular}
\caption{Evaluation results of Few-shot ($FS$) ICL two-stage ranking demonstration examples selection using BM-25 with random ordering for top-k most similar examples (one-shot per class) for Olmo-7b-Instruct, Gemma-7b-it, Llama3-8b-Instruct on HS-Brexit, MD-Agreement and ConvAbuse datasets. $RL$ denotes role-playing, $aggr$ aggregated and $disaggr$ disaggregated labels. The best JSD scores are highlighted in \textbf{bold}, and the best F1 and CE scores are \underline{underlined}.}
\label{tab:results_bm25_re_rank}
\end{table}

\clearpage

\renewcommand{\arraystretch}{0.5} 
\setlength{\tabcolsep}{1pt} 
\begin{table}[!ht]
\centering
\tiny
\begin{tabular}{lllcccc}
    \toprule
    \textbf{Dataset} & \textbf{LLM} & \textbf{Approach} & \textbf{Acc↑} & \textbf{F1↑} & \textbf{JSD↓} & \textbf{CE↓}\\
    \midrule
    \multirow{18}{*}{HS-Brexit} 
    & \multirow{3}{*}{Olmo-7b-Instruct}      & Baseline\_aggr\_FS & 57.14 & \underline{46.82} & \textbf{0.35} & 0.93  \\
 &   & Baseline\_aggr\_FS\_CL & 51.19 & 42.90 & 0.38 & 1.03   \\
    &                                       & Baseline\_aggr\_FS\_RL & 56.23 & 43.87 & 0.37 & 0.97  \\
     &                                       & Baseline\_aggr\_FS\_RL\_CL & 50.75 & 41.67 & 0.39 & 1.14  \\
    &                                       & MultiP\_aggr\_hard FS  & 54.76 & \underline{45.89} & \textbf{0.34} & \underline{0.85}      \\
    &    & MultiP\_aggr\_hard FS\_CL & 50.59 & 43.09 & 0.36 & 0.92    \\
     &                                       & MultiP\_aggr\_hard FS\_RL  & 54.16 & 44.86 & \textbf{0.35} & 0.94  \\
     &                                       & MultiP\_aggr\_hard FS\_RL\_CL      & 48.21 & 41.47 & 0.38 & 1.03 \\
       &                                       & MultiP\_disaggr\_soft  FS & 42.26 & 33.49 &  0.46 & 5.29    \\
       &                                       & MultiP\_disaggr\_soft FS\_CL & 45.23 & 36.60 & 0.45 & 5.29     \\
      &                                       & MultiP\_disaggr\_soft FS\_RL  & 42.85 & 35.71 & 0.47 & 5.97      \\
      &                                       & MultiP\_disaggr\_soft  FS\_RL\_CL    & 46.42 & 38.59 &0.45 & 5.03    \\
       &                                       & MultiP\_disaggr\_hard FS   &  39.28 & 35.23 & 0.52 & 11.02\\
       &                                       & MultiP\_disaggr\_hard FS\_CL      &  38.09 & 34.75 & 0.54 & 11.24\\
        &                                       & MultiP\_disaggr\_hard FS\_RL      & 25.59 & 25.14 & 0.61 & 14.11 \\
        &                                       & MultiP\_disaggr\_hard FS\_RL\_CL      & 19.64 & 19.61 & 0.64 & 15.35  \\
    \cmidrule(lr){2-7}
    & \multirow{3}{*}{Gemma-7b-it}          & Baseline\_aggr\_FS & 78.57 & \underline{51.02} & 0.32 & 0.58   \\
 &   & Baseline\_aggr\_FS\_CL  & 76.19 & 49.62 & 0.31 & 0.57       \\
    &                                       & Baseline\_aggr\_FS\_RL  & 73.28 & 45.76 & 0.37 & 0.67  \\
     &                                       & Baseline\_aggr\_FS\_RL\_CL & 76.24 & 48.63 & 0.35 & 0.62  \\
    &                                       & MultiP\_aggr\_hard FS & 69.64 & 50.18 & \textbf{0.28} & 0.52       \\
    &                                       & MultiP\_aggr\_hard FS\_CL & 76.78 & \underline{51.76} & \textbf{0.27} & \underline{0.50}       \\
     &                                       & MultiP\_aggr\_hard FS\_RL & 58.92 & 45.66 & 0.35 & 0.68       \\
     &                                       & MultiP\_aggr\_hard FS\_RL\_CL  & 60.71 & 46.76 & 0.34 & 0.66  \\
       &                                       & MultiP\_disaggr\_soft  FS  & 54.76 & 47.62 & 0.43 & 5.48      \\
       &                                       & MultiP\_disaggr\_soft FS\_CL & 57.73 & 49.74 & 0.40 & 4.01       \\
      &                                       & MultiP\_disaggr\_soft FS\_RL  & 56.54 & 48.89 & 0.42 & 5.59      \\
      &                                       & MultiP\_disaggr\_soft  FS\_RL\_CL & 60.71 & \underline{51.87} & 0.39 & 4.01       \\
       &                                       & MultiP\_disaggr\_hard FS      & 17.85 & 17.81 & 0.69 & 16.14 \\
       &                                       & MultiP\_disaggr\_hard FS\_CL      & 17.26 & 17.23 & 0.68 & 15.95 \\
        &                                       & MultiP\_disaggr\_hard FS\_RL      & 19.64 & 19.64 & 0.67 & 15.43 \\
        &                                       & MultiP\_disaggr\_hard FS\_RL\_CL  & 16.67 & 16.62 &  0.69 & 16.61 \\
    \cmidrule(lr){2-7}
    & \multirow{3}{*}{Llama3-8b-Instruct}   & Baseline\_aggr\_FS   & 58.92 & 51.16 & \textbf{0.34} & 1.53        \\
 &   & Baseline\_aggr\_FS\_CL & 58.33 & 51.25 & \textbf{0.35} & 1.66         \\
    &                                       & Baseline\_aggr\_FS\_RL  &  55.39 & 47.63 & 0.37 & 1.76 \\
     &                                       & Baseline\_aggr\_FS\_RL\_CL &  56.70 & 48.04 & 0.38 & 1.79 \\
    &                                       & MultiP\_aggr\_hard FS & 56.54 & 49.54 & 0.36 & \underline{1.38}       \\
    &                                       & MultiP\_aggr\_hard FS\_CL & 52.38 & 46.89 & 0.36 & 1.47        \\
     &                                       & MultiP\_aggr\_hard FS\_RL & 55.35 & 49.07 & 0.37 & 1.61       \\
     &                                       & MultiP\_aggr\_hard FS\_RL\_CL & 49.40 & 44.70 & 0.42 & 1.90  \\
       &                                       & MultiP\_disaggr\_soft  FS & 59.52 & 52.14 & 0.37 & 1.99       \\
       &                                       & MultiP\_disaggr\_soft FS\_CL & 54.17 & 48.20 & 0.39 & 2.48      \\
      &                                       & MultiP\_disaggr\_soft FS\_RL & 55.35 & 49.07 & 0.38 & 2.56      \\
      &                                       & MultiP\_disaggr\_soft  FS\_RL\_CL & 46.42 & 42.48 & 0.42 & 3.17       \\
       &                                       & MultiP\_disaggr\_hard FS      & 63.69 & \underline{54.03} & \textbf{0.35} & 6.13 \\
       &                                       & MultiP\_disaggr\_hard FS\_CL      &  61.31 & 52.89 & 0.37 & 6.72\\
        &                                       & MultiP\_disaggr\_hard FS\_RL & 26.19 & 26.02 & 0.63 & 14.71  \\
        &                                       & MultiP\_disaggr\_hard FS\_RL\_CL      & 28.57 & 28.21 & 0.61 & 14.27  \\
    \midrule
    \multirow{18}{*}{MD-Agr} 
    & \multirow{3}{*}{Olmo-7b-Instruct}      & Baseline\_aggr\_FS        & 59.47 & 48.79 & 0.28 & 0.42  \\
 &   & Baseline\_aggr\_FS\_CL        & 61.23 & 48.91 & 0.28 & 0.44  \\
    &                                       & Baseline\_aggr\_FS\_RL      & 57.89 & \underline{57.51} & \textbf{0.26} & 0.51  \\
     &                                       & Baseline\_aggr\_FS\_RL\_CL      & 58.11 & \underline{57.55} & 0.30 & 0.51  \\
    &                                       & MultiP\_aggr\_hard FS      & 66.50 & 40.67 & \textbf{0.26} & \underline{0.37}  \\
    &                                       & MultiP\_aggr\_hard FS\_CL      & 67.20 & 41.27 & \textbf{0.27} & \underline{0.38}  \\
     &                                       & MultiP\_aggr\_hard FS\_RL      & 65.39 & 42.96 & \textbf{0.26} & \underline{0.38}  \\
     &                                       & MultiP\_aggr\_hard FS\_RL\_CL      & 66.46 & 43.11 & \textbf{0.26} & \underline{0.38}  \\
       &                                       & MultiP\_disaggr\_soft  FS      & 50.81 & 47.43 & \textbf{0.27} & 1.35  \\
       &                                       & MultiP\_disaggr\_soft FS\_CL      & 50.92 & 47.72 & 0.28 & 1.35  \\
      &                                       & MultiP\_disaggr\_soft FS\_RL      & 61.01 & 50.98 & 0.32 & 1.51  \\
      &                                       & MultiP\_disaggr\_soft  FS\_RL\_CL      & 61.59 & 51.23 & 0.33 & 1.50  \\
       &                                       & MultiP\_disaggr\_hard FS      & 48.45 & 46.83 & 0.63 & 3.29 \\
       &                                       & MultiP\_disaggr\_hard FS\_CL      & 48.85 & 47.13 & 0.62 & 3.27 \\
        &                                       & MultiP\_disaggr\_hard FS\_RL      & 51.17 & 48.72 & 0.59 & 3.15  \\
        &                                       & MultiP\_disaggr\_hard FS\_RL\_CL      & 51.46 & 48.92 & 0.59 & 3.14  \\

    \cmidrule(lr){2-7}
    & \multirow{3}{*}{Gemma-7b-it}          & Baseline\_aggr\_FS        & 67.12  & 60.42 & 0.45  & 0.56  \\
    & & Baseline\_aggr\_FS\_CL         & 68.02  & 61.31 & 0.46  & \underline{0.54}  \\
    &                                       & Baseline\_aggr\_FS\_RL      & 65.78 & 41.56 & 0.42 & \underline{0.53}  \\
    &                                       & Baseline\_aggr\_FS\_RL\_CL      & 65.78 & 41.58 & 0.41 & \underline{0.53}  \\
    &                                       & MultiP\_aggr\_hardFS      & 66.27 & 60.20 & \textbf{0.40} & 0.62  \\
    &                                       & MultiP\_aggr\_hard FS\_CL      & 66.59 & 60.80 & \textbf{0.39} & 0.62  \\
    &                                       & MultiP\_aggr\_hard FS\_RL      & 66.86 & \underline{64.36} & 0.42 & 0.71  \\
     &                                       & MultiP\_aggr\_hard FS\_RL\_CL      & 65.26 & \underline{64.86} & 0.41 & 0.71  \\
     &                                       & MultiP\_disaggr\_soft  FS     & 58.66 & 53.71 & 0.50 & 2.27  \\
     &                                       & MultiP\_disaggr\_soft FS\_CL     & 58.66 & 53.76 & 0.50 & 2.29  \\
     &                                       & MultiP\_disaggr\_soft FS\_RL     & 59.27 & 55.68 & 0.49 & 2.53  \\
     &                                       & MultiP\_disaggr\_soft FS\_RL\_CL    & 60.07 & 55.98 & 0.49 & 2.54  \\
      &                                       & MultiP\_disaggr\_hard FS     & 50.39 & 45.68 & 0.62 & 5.90  \\
       &                                       & MultiP\_disaggr\_hard FS\_CL     & 51.32 & 46.68 & 0.61 & 5.87  \\
       &                                       & MultiP\_disaggr\_hard FS\_RL     & 52.43 & 46.25 & 0.57 & 5.59  \\
       &                                       & MultiP\_disaggr\_hard FS\_RL\_CL     & 52.63 & 46.65 & 0.56 & 5.59  \\
    \cmidrule(lr){2-7}
    & \multirow{3}{*}{Llama3-8b-Instruct}   & Baseline\_aggr\_FS         & 53.48 & 53.05 & 0.30 & 0.54  \\
    & & Baseline\_aggr\_FS\_CL         & 54.18 & 53.81 & 0.28 & 0.52  \\
    &                                       & Baseline\_aggr\_FS\_RL      & 52.96 & 52.83 & 0.30 & 0.54  \\
    &                                       & Baseline\_aggr\_FS\_CL      & 53.26 & 53.13 & 0.27 & 0.50  \\
    &                                       & MultiP\_aggr\_hard FS      & 59.99 & \underline{59.24} & 0.29 & 0.43  \\
    &                                       & MultiP\_aggr\_hard FS\_CL      & 59.99 & \underline{59.11} & 0.28 & 0.42  \\
     &                                       & MultiP\_aggr\_hard FS\_RL      & 59.37 & \underline{59.22} & \textbf{0.26} & \underline{0.35}  \\
     &                                       & MultiP\_aggr\_hard FS\_RL\_CL      & 59.79 & \underline{59.52} & \textbf{0.25} & \underline{0.35}  \\
       &                                       & MultiP\_disaggr\_soft  FS     & 52.69 & 48.70 & 0.47 & 1.79  \\
       &                                       & MultiP\_disaggr\_soft FS\_CL     & 52.69 & 48.85 & 0.47 & 1.78  \\
        &                                       & MultiP\_disaggr\_soft FS\_RL     & 57.28 & 53.16 & 0.48 & 1.76  \\
        &                                       & MultiP\_disaggr\_soft FS\_RL\_CL     & 57.26 & 52.98 & 0.48 & 1.75  \\
        &                                       & MultiP\_disaggr\_hard FS     & 49.76 & 45.32 & 0.69 & 4.81  \\
         &                                       & MultiP\_disaggr\_hard FS\_CL     & 49.82 & 45.74 & 0.68 & 4.79  \\
          &                                       & MultiP\_disaggr\_hard FS\_RL     & 50.49 & 47.96 & 0.57 & 4.12  \\
          &                                       & MultiP\_disaggr\_hard FS\_RL\_CL     & 50.63 & 48.02 & 0.56 & 4.10  \\
          \bottomrule
\end{tabular}
\caption{Evaluation results of Few-shot ($FS$) ICL BM-25 demonstration example selection (one-shot per class) with random ordering and with curriculum learning for Olmo-7b-Instruct, Gemma-7b-it, Llama3-8b-Instruct on HS-Brexit, MD-Agreement datasets. $RL$ denotes role-playing, $CL$ curriculum learning (if not explicitly stated, the ordering is random), $aggr$ aggregated and $disaggr$ disaggregated labels. The best JSD scores are highlighted in \textbf{bold}, and the best F1 and CE scores are \underline{underlined}.}
\label{tab:results_bm25_hs_md}
\end{table} 

\renewcommand{\arraystretch}{0.5} 
\setlength{\tabcolsep}{1pt} 
\begin{table}[!ht]
\centering
\tiny
\begin{tabular}{lllcccc}
    \toprule
    \textbf{Dataset} & \textbf{LLM} & \textbf{Approach} & \textbf{Acc↑} & \textbf{F1↑} & \textbf{JSD↓} & \textbf{CE↓} \\
    \midrule
    \multirow{18}{*}{ConvAbuse} 
    & \multirow{3}{*}{Olmo-7b-Instruct}      & Baseline\_aggr\_FS        & 75.71 & 52.99 & \textbf{0.15} & \underline{0.26}  \\
    &  & Baseline\_aggr\_FS\_CL         & 75.95 & 53.27 & \textbf{0.14} & \underline{0.25}  \\
    &                                       & Baseline\_aggr\_FS\_RL      & 71.19 & 46.61 & 0.21 & 0.34  \\
    &                                       & Baseline\_aggr\_FS\_RL\_CL      & 71.62 & 47.14 & 0.20 & 0.33  \\
    &                                       & MultiP\_aggr\_hard FS      & 70.95 & 45.94 & 0.16 & 0.31  \\
    &                                       & MultiP\_aggr\_hard FS\_CL      & 71.17 & 46.01 & \textbf{0.15} & 0.30  \\
    &                                       & MultiP\_aggr\_hard FS\_RL      & 71.27 & 46.32 & \textbf{0.15} & 0.31  \\
    &                                       & MultiP\_aggr\_hard FS\_RL\_CL      & 58.33 & \underline{55.44} & \textbf{0.14} & 0.44  \\
     &                                       & MultiP\_disaggr\_soft  FS      & 57.89 & 51.89 & 0.18 & 0.45  \\
     &                                       & MultiP\_disaggr\_soft FS\_CL      & 57.89 & 51.95 & 0.18 & 0.44  \\
       &                                       & MultiP\_disaggr\_soft  FS\_RL      & 61.51 & 53.27 & 0.23 & 0.97  \\
         &                                       & MultiP\_disaggr\_soft FS\_RL\_CL      & 61.64 & 53.29 & 0.22 & 0.96  \\
     &                                       & MultiP\_disaggr\_hard FS      & 49.28 & 42.24 & 0.23 & 0.93  \\
      &                                       & MultiP\_disaggr\_hard FS\_CL      & 49.45 & 42.51 & 0.22 & 0.93  \\
      &                                       & MultiP\_disaggr\_hard FS\_RL      & 49.44 & 43.93 & 0.21 & 0.45  \\
      &                                       & MultiP\_disaggr\_hardFS\_RL\_CL      & 49.89 & 45.05 & 0.19 & 0.45  \\
    \cmidrule(lr){2-7}
    & \multirow{3}{*}{Gemma-7b-it}          & Baseline\_aggr\_FS        & 75.71  & 52.99 & \textbf{0.15}  & \underline{0.29}  \\
    & & Baseline\_aggr\_FS\_CL         & 76.21  & 53.38 & \textbf{0.14}  & 0.30  \\
    &                                       & Baseline\_aggr\_FS\_RL      & 81.19 & 46.61 & 0.21 & 0.32  \\
    &                                       & Baseline\_aggr\_FS\_RL\_CL      & 81.19 & 46.73 & 0.20 & 0.32  \\
    &                                       & MultiP\_aggr\_hard FS      & 80.95 & 45.94 & \textbf{0.15} & \underline{0.28}  \\
    &                                       & MultiP\_aggr\_hard FS\_CL      & 80.95 & 45.96 & \textbf{0.14} & \underline{0.28}  \\
    &                                       & MultiP\_aggr\_hard FS\_RL      & 57.38 & \underline{59.05} & 0.20 & 0.34  \\
    &                                       & MultiP\_aggr\_hard FS\_RL\_CL      & 57.72 & \underline{59.24} & 0.19 & 0.34  \\
     &                                       & MultiP\_disaggr\_soft  FS     & 63.55 & 54.89 & 0.19 & 0.43  \\
     &                                       & MultiP\_disaggr\_soft FS\_RL     & 63.60 & 54.92 & 0.19 & 0.41  \\
     &                                       & MultiP\_disaggr\_soft FS\_RL\_CL     & 50.11 & 49.21 & 0.28 & 0.39  \\
      &                                       & MultiP\_disaggr\_hard FS     & 51.32 & 46.94 & 0.56 & 4.09  \\
      &                                       & MultiP\_disaggr\_hardFS\_CL     & 51.62 & 47.10 & 0.54 & 4.06  \\
       &                                       & MultiP\_disaggr\_hardFS\_RL     & 55.71 & 49.06 & 0.61 & 4.36  \\
       &                                       & MultiP\_disaggr\_hard FS\_RL\_CL     & 55.69 & 49.12 & 0.59 & 4.32  \\
    \cmidrule(lr){2-7}
    & \multirow{3}{*}{Llama3-8b-Instruct}   & Baseline\_aggr\_FS         & 72.50 & 64.17 & 0.42 & 0.54  \\
    & & Baseline\_aggr\_FS\_CL         & 73.26 & 64.89 & 0.41 & 0.52  \\
    &                                       & Baseline FS\_RL      & 63.80 & 53.33 & \textbf{0.32} & 0.46  \\
    &                                       & Baseline FS\_RL\_CL      & 64.80 & 53.33 & \textbf{0.32} & 0.46  \\
    &                                       & MultiP\_aggr\_hard FS      & 77.26 & \underline{66.57} & 0.41 & 0.53  \\
     &                                       & MultiP\_aggr\_hard FS\_CL      & 77.66 & \underline{67.17} & 0.40 & 0.51  \\
     &                                       &MultiP\_aggr\_hard FS\_RL      & 62.26 & 56.70 & \textbf{0.32} & \underline{0.45}  \\
     &                                       & MultiP\_aggr\_hard FS\_RL\_CL      & 62.76 & 57.30 & \textbf{0.31} & \underline{0.44}  \\
      &                                       & MultiP\_disaggr\_soft  FS      & 54.14 & 53.09 & 0.39 & 2.27  \\
      &                                       & MultiP\_disaggr\_soft FS\_CL      & 54.76 & 53.45 & 0.37 & 2.23  \\
       &                                       & MultiP\_disaggr\_soft FS\_RL      & 57.63 & 56.45 & 0.37 & 0.56  \\
       &                                       & MultiP\_disaggr\_soft  FS\_RL\_CL      & 57.63 & 56.53 & 0.36 & 0.56  \\
        &                                       & MultiP\_disaggr\_hard FS      & 51.24 & 43.94 & 0.41 & 5.39  \\
         &                                       & MultiP\_disaggr\_hardFS\_CL      & 51.27 & 44.12 & 0.40 & 5.39  \\
        &                                       & MultiP\_disaggr\_hard FS\_RL      & 48.12 & 44.44 & 0.57 & 6.12  \\
        &                                       & MultiP\_disaggr\_hard FS\_RL\_CL      & 48.12 & 44.53 & 0.56 & 6.10  \\
        \bottomrule
\end{tabular}
\caption{Evaluation results of Few-shot ($FS$) ICL BM-25 demonstration example selection (one-shot per class) with random ordering and with curriculum learning for Olmo-7b-Instruct, Gemma-7b-it, Llama3-8b-Instruct on ConvAbuse dataset. $RL$ denotes role-playing, $CL$ curriculum learning (if not explicitly stated, the ordering is random), $aggr$ aggregated and $disaggr$ disaggregated labels. The best JSD scores are highlighted in \textbf{bold}, and the best F1 and CE scores are \underline{underlined}.}
\label{tab:results_bm25_conv}
\end{table}


\end{document}